**Deep learning assisted high resolution microscopy image processing for phase segmentation in functional composite materials**

*Ganesh Raghavendran[a], Bing Han[a], Fortune Adekogbe[d], Shuang Bai[a], Bingyu Lu[a], William Wu[e], Minghao Zhang[c,*], Ying Shirley Meng[a,c,*]*

a Department of NanoEngineering, University of California San Diego, La Jolla, CA 92093

b Materials Science and Engineering, University of California San Diego, La Jolla, CA 92093

c Pritzker School of Molecular Engineering, University of Chicago, Chicago, IL 60637

d Department of Chemical and Petroleum Engineering, University of Lagos, Lagos, Nigeria 101017

e Department of Computer Science, University of California San Diego, La Jolla, CA 92093

Correspondence to miz016@uchicago.edu; shirleymeng@uchicago.edu

## ■ ABSTRACT

In the domain of battery research, the processing of high-resolution microscopy images is a challenging task, as it involves dealing with complex images and requires a prior understanding of the components involved. The utilization of deep learning methodologies for image analysis has attracted considerable interest in recent years, with multiple investigations employing such techniques for image segmentation and analysis within the realm of battery research. However, the automated analysis of high-resolution microscopy images for detecting phases and components in composite materials is still an underexplored area. This work proposes a novel workflow for FFT-based segmentation, periodic component detection and phase segmentation from raw high resolution Transmission Electron Microscopy (TEM) images using a trained U-Net segmentation model. The developed model can expedite the detection of components and their phase segmentation, diminishing the temporal and cognitive demands associated with scrutinizing an extensive array of TEM images, thereby mitigating the potential for human errors. This approach presents a novel and efficient image analysis approach with broad applicability beyond the battery field and holds potential for application in other related domains characterized by phase and composition distribution, such as alloy production.

## ■ INTRODUCTION

In the era of information, the emphasis has transitioned from data collection to data processing. Currently, an abundance of data is accessible across various research domains, and the primary hurdle lies in extracting meaningful information through its processing. This principle holds true within the realm of materials science, where researchers endeavor to derive valuable insights from their experimental samples, using high resolution microscopy imaging [1]. The post-processing of these images has always been a challenge, owing to the complexity of the high-resolution image and the need for prior knowledge of components involved. Additionally, a comprehensive examination necessitates the acquisition of multiple images of the same sample, thereby augmenting the burden of post-processing.

In recent years, high-resolution imaging has become an indispensable tool in the field of battery material research, playing a vital role in the development of strategies to address the increasing energy demand. High-Resolution Transmission Electron Microscopy (HRTEM) stands out as a powerful technique used for investigating the microstructure of battery materials, encompassing cathodes, anodes, and electrolytes, with atomic resolution [2]. Among the various battery systems under investigation, Li metal batteries have received significant attention due to their propensity to store at least 33% more power per pound than traditional Li-ion batteries, rendering them suitable for a wide range of applications, including electric vehicles, renewable energy integration, and grid-scale energy storage [3]. It is important to note that cycled Li metal, due to presence of oxides, carbonates, and sulfides, exhibits sensitivity to electron beam. Consequently, cryogenic Electron Microscopy (cryo-EM) technology has emerged, expanding the feasibility and necessity of high-resolution imaging in battery research, particularly for studying electron beam-sensitive anode materials like electrochemically cycled Li metal [4,5]. Solid Electrolyte Interphase (SEI) formed in the anode materials after cycling is regarded as one of the least understood systems in the battery community [6,7]. Studying the SEI is of scientific interest due to its crucial role in enabling long-term cycling in battery systems, as well as its complex composition of both organic and inorganic compounds. HRTEM, particularly when used in conjunction with cryogenic techniques, enables the elucidation of the components of the thin SEI layer, which typically has a thickness in nanometer scale [8]. Such characterization methods facilitate SEI engineering and the development of improved electrolyte systems.

In addition to the arduous task of sample preparation for HRTEM, processing the resulting images can also present significant challenges due to the high resolution and large amount of data produced [9]. Furthermore, *in situ* TEM studies targeting the dynamic interplay of properties, structures, and compositions within nanostructures yield substantial datasets acquired at elevated frame rates, posing an exceedingly formidable task for comprehensive data analysis [10]. Recently, novel image analysis techniques have emerged within the framework of deep learning, a data processing approach that has gained tremendous popularity over the past decade. The proliferation of deep learning techniques within the field of image analysis has been spurred by several factors, such as the expanded availability of labeled datasets of significant size, notable advances in the realm of deep learning research, and the emergence of powerful high-performance frameworks like PyTorch and TensorFlow [11,12].

In the field of battery research, several studies have employed machine learning and deep learning techniques for image analysis [13–15]. The image analysis investigations have largely focused on segmenting microscopy and tomography images of electrode materials [16,17]. Automated analysis of high-resolution microscopy images for detection of phases and components in composite battery materials is an area that remains relatively unexplored. In contrast, there have been several notable works in materials science that have focused on detecting phases and components from HRTEM images. For instance, Liu et al. employed an unsupervised clustering algorithm combined with a scanning window technique to detect and group different phases from a TEM image of an AM Inconel 718 alloy [18]. The technique can detect and map multiple phases, but there is a possibility of the same phase getting identified more than once and the chemical composition of the phases is not identified. While the method exhibits efficiency in phase mapping, its suitability for analyzing battery materials is more limited compared to alloys. Zhang et al. proposed an improved Local Contrast Attention – UNet (LCA-UNet) for detecting Fast Fourier Transform (FFT) features from TEM images of Zirconium oxides [19]. The authors utilized a scanning window technique to generate the FFT, then identified the features and labeled the corresponding phase window. The study used a 64 x 64-pixels region of the 256 x 256-pixels window for detection, potentially losing information outside the chosen region. The employed method disregarded the inclusion of asymmetrical characteristics, potentially leading to the loss of pertinent information, and the designated window was a square region rather than accurately representing the intended area. Furthermore, the intricate composition of the cycled Li metal anode samples may lead to low-contrast images that are challenging to process. In fact, using smaller-scale 2-D FFTs may exacerbate this issue, even with advanced deep learning models such as LCA-UNet.

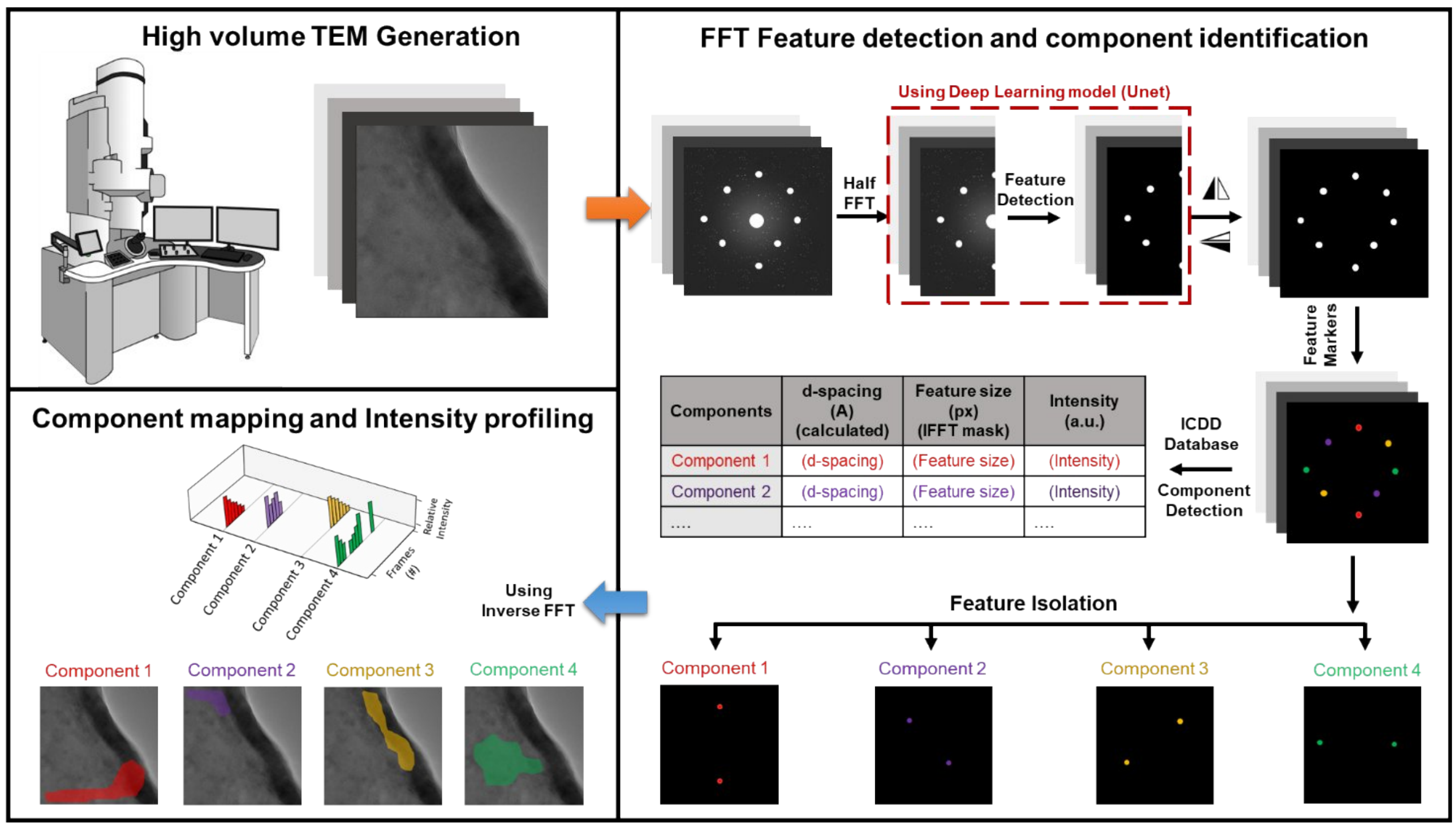

**Figure 1.** Schematic of the program flow – High volume of TEM data is acquired for analysis → FFT features are detected using deep learning model → Components found out from detected features using ICDD (International Centre for Diffraction Data) database → Each feature is isolated and mapped using IFFT → Component detection program is used for high volume TEM processing → Intensity profiling and mapping of components obtained.

The workflow in this study for FFT-based segmentation, periodic component detection and phase segmentation from HRTEM images of battery materials is presented in **Fig. 1**. First, a FFT diffractogram is generated from the HRTEM image, and a trained U-Net segmentation model is used to detect features from the generated 2-D FFT cropped half (1024 x 512 pixels). Exploiting the innate symmetry present in the 2-D FFTs, half cropped 2-D FFTs (1024 x 512 pixels) were used for model training. The half cropped 2-D FFTs performed better than the full 2-D FFTs (1024 x 1024 pixels). The d-spacing of the identified features from model generated 2-D FFT is calculated and compared with the material database to detect the components present in the TEM image. Finally, the inverse FFT (IFFT) image is generated for each component by masking the actual 2-D FFT using the deep learning model, and watershed segmentation is used to map the periodic components present in the TEM image. For high-volume TEM processing, this process is repeated for all the TEM images to be analyzed, and the intensity profile for each component is generated to observe their evolution over the imaging period. The developed segmentation model can expedite component detection and phase segmentation, not only in the battery field but also in alloy manufacturing. The qualitative and quantitative analysis of TEM images in this work decreases the time and effort necessary for evaluating substantial quantities of TEM images while mitigating the potential for human error caused by cognitive task-induced state fatigue [20] of analyzing a large set of TEM images (e.g ~100 images).

## ■ MATERIALS AND METHODS

### TEM Sample Preparation:

LiF powder was ground in an Ar gas glove box and the ground LiF powder was dispersed on the TEM Cu grid. The TEM Cu grid is loaded to Melbuild vacuum transfer holder under Ar atmosphere and the sample is transferred into the TEM column without any air exposure. The experiment is performed at room temperature for beam damage analysis. For cryo temperature measurement, liquid nitrogen was added to the dewar when the holder is fully inserted into the column. The system stabilized at ~-165℃ for cryo temperature beam damage analysis.

### TEM Imaging:

We used a ThermoFisher Talos F200X TEM electron microscope system with super-low-dose TEM techniques to characterize samples. The low dose HRTEM images were acquired with controllable electron dose rate (50-1000 e $A^{-2}$ $s^{-1}$) at FEI Ceta 16M camera and low dose system. The pixel size of the TEM image corresponds to 0.037 nm/pixel.

**FFT generation from TEM images**

TEM images can contain complex structures with multiple crystallographic and amorphous phases. Therefore, extracting and interpreting statistical information and uncovering the underlying physical mechanisms can pose significant challenges. FFT is useful for identifying periodic patterns in TEM images, such as lattice fringes, and for extracting information about the crystallographic orientation and symmetry of the sample [21]. For this study, the Cryo-EM images of cycled Li metal anode in different electrolytes are used to introduce variability and to reduce redundancy. Images of magnification >300 kx are chosen for getting SEI layer of the anode in better resolution. The Cryo-EM images of format .dm3/.dm4 generated by GATAN micrograph software are used specifically. The TEM image is generated from the .dm3/.dm4 files using the DM3lib python library [22] and the 2-D FFT is generated from the TEM image using the FFT function from the OpenCV2 library in Python. The mathematical representation of the 2D Discrete Fourier Transform (DFT) for an image signal $f(x,y)$, across an x-y plane is:

$$F(u,v) = \frac{1}{M * N} \sum_{x=0}^{M-1} \sum_{y=0}^{N-1} f(x,y)\, e^{-2\pi i\left(\frac{ux}{M}+\frac{vy}{N}\right)} \quad \#(1)$$

, where $F(u,v)$ is the Discrete Fourier Transform of $f(x,y)$, $u$ and $v$ are the spatial frequencies in the $x$ and $y$ directions, respectively, and M, N are their associated sizes. The sum is taken over the entire $x$–$y$ plane. In the field of image processing, the Fourier transform is commonly employed to analyze and manipulate the frequency components of an image. To achieve a more visually meaningful representation, the zero-frequency component is shifted to the center of the transformed array, and the resulting complex number is converted into an absolute value. A logarithmic operation is then applied to the absolute value for perceptual scaling [23]. This transformed matrix is subsequently converted into an intensity image containing only the normalized logarithm of the absolute values of the complex numbers, wherein the pixel values range from 0 (representing black) to 1 (representing white). The resulting image is then cropped from the center to generate a reduced 2-D FFT with a size of 2048 x 2048 pixels.

**FFT preprocessing**

GATAN [24] software-generated FFTs are used as a reference for pre-processing training 2-D FFTs, as their manual display adjustments allow for effective noise reduction and feature enhancement. However, manual adjustment is not feasible for large datasets. Therefore, an automated Python-based denoising method was developed.

Analysis of GATAN's display controls (**Fig. S2(a)**) reveals that suppressing pixel intensities below the histogram's mean value significantly improves feature visibility (marked by dashed red line). This observation inspired the Python denoising process. First, the mean pixel value of the noisy FFT generated using Python (**Fig. S2(b)**) is calculated. This mean is then subtracted from every pixel in the image and the minimum value is clipped to 0, effectively removing negative values. This process, similar to GATAN's manual adjustment, reduces the background noise but has poor feature contrast (**Fig. S2(c)**).

To further improve visual similarity with the GATAN FFTs, the brightness and contrast of the denoised image are adjusted. This step involves linear scaling using the OpenCV library's convertScaleAbs function. Scaling parameters (alpha = 3 for contrast, beta = 11 for brightness) that maximize similarity with the GATAN output were empirically determined (**Fig. S2(d)**).

To quantify the similarity between the processed FFTs and the reference GATAN FFT, the Structural Similarity Index Measure (SSIM) metric is used. SSIM [25], implemented using the scikit-image library in Python, is chosen for its ability to assess perceived similarity, which aligns with the goal of generating FFTs that are visually similar to the human-adjusted GATAN outputs. SSIM considers luminance, contrast, and structure, making it more suitable than pixel-based metrics like Mean Squared Error (MSE) which are less sensitive to structural changes [26] (**Fig. S2**).

A positive correlation between SSIM scores and the performance metrics of models trained on different FFT datasets (noisy, denoised, enhanced) was observed. This correlation, detailed in the ablation study section (**Table S5**), highlights the importance of effective FFT denoising for optimal model training.

**Component detection from FFT– Radial profiling**

Radial profiling is a widely used and straightforward approach for analyzing 2-D FFTs [27] . To perform radial profiling, the distance of each pixel from the center point is first calculated using the Euclidean distance formula:

$$r = \sqrt{(x - x_0)^2 + (y - y_0)^2} \# (2)$$

Here, $r$ is the radial distance, $(x,y)$ are coordinates of the pixel and $(x_0,y_0)$ are the coordinates of the center of the 2-D FFT. Then, the pixels that are equidistant from the center are binned together, and the average intensity within each bin is computed. This results in a one-dimensional profile representing the average intensity as a function of radial distance.

The resulting integrated values are plotted against the real distance of each circular band, which is determined by calculating the reciprocal distance from the center of the 2-D FFT (**Fig. S3 (g-i)**).

To convert the reciprocal distance to real distance, the pixel size information of the TEM image file (.dm3/.dm4) is extracted using DM3lib and applied to the following formula:

$$real\ distance = \frac{pixelsize * N}{r} \# (3)$$

Here, *pixelsize* is the calibration information obtained from the metadata of the TEM file that indicates the real size of a pixel in the TEM image, *N* is the corresponding size of the TEM image, *r* is the radial distance of the pixel in the FFT calculated from equation [2]. The periodic components present in the TEM image can be identified by the peaks obtained from the diffraction-like graph generated using the circular integration method. The position of the peak corresponds to the d-spacing of the periodic components, aiding in the identification of the components present in the image. However, the analysis of the noisy 2-D FFTs using the circular integration method results in a diffraction-like graph that is also noisy, making it difficult to detect the peaks accurately (**Fig. S3(g-i, j)**).

**Automated FFT training set generation:**

The preprocessing of 2-D FFTs generated from HRTEM images (detailed in previous section) for the purpose of training a segmentation model involves the use of Gaussian filters, which is a standard image processing technique available in the widely used OpenCV python package [28]. The Gaussian filter was selected over other common denoising approaches (e.g., median, bilateral) due to its ability to preserve edge information while high-frequency noise is effectively suppressed—characteristics particularly important for 2-D FFTs. 2-D FFTs were chosen for filter application rather than raw TEM images because the feature detection algorithm operates in the frequency domain, where periodic features can be isolated and background noise can be reduced through Gaussian filtering. Additionally, direct application of Gaussian blur on TEM images could potentially smudge or blur the sharp lattice fringes, which would result in significant loss of critical crystallographic information. By applying the filtering in the frequency domain instead, the integrity of the original lattice structures can be preserved while still improving signal quality for subsequent analysis.

A comprehensive parametric study was conducted comparing different Gaussian kernel sizes ((1,1), (3,3), and (5,5)) against unfiltered images ("No Blur") across various feature sizes (7-40 pixels). Each filter configuration was quantitatively evaluated using Signal-to-Noise Ratio (SNR) and Contrast-to-Noise Ratio (CNR) metrics (**Fig. S4**). SNR and CNR were calculated using binary masks derived from labeled images to identify signal (feature – labelled white) and noise (background – labelled black) regions. The SNR was computed as 20 times the logarithm (base 10) of the ratio between the mean signal intensity ($\mu_{signal}$) and the standard deviation of the noise ($\sigma_{noise}$), expressed in decibels (dB).

$$SNR = 20 * \log_{10}\left(\frac{\mu_{signal}}{\sigma_{noise}}\right) \# (4)$$

The CNR was determined by calculating the difference between the mean signal intensity ($\mu_{signal}$) and mean noise intensity ($\mu_{noise}$),divided by the standard deviation of the noise ($\sigma_{noise}$).

$$CNR = \frac{\mu_{signal} - \mu_{noise}}{\sigma_{noise}} \# (5)$$

These metrics quantitatively assess the quality of feature detection, where higher SNR values indicate better signal clarity and higher CNR values indicate better feature distinguishability from the background noise.

Superior performance was demonstrated by the Gaussian filter with kernel size (3,3), with the highest average SNR (21.03 ± 1.92 dB) and CNR (7.83 ± 3.35) values being yielded across all object sizes. Optimal enhancement for smaller features (7-22px) was provided by the (3,3) kernel, while slightly better results for larger features (31-40px) were shown with (5,5). However, since features across all size ranges are contained in the dataset, (3,3) was selected as the best compromise (**Fig. S5)**.

In contrast, the production of labeled data involves manual input from experts in the field who utilize Adobe Photoshop software to identify and mark features in the 2-D FFTs.

To generate a training set for the proposed deep learning-based framework for FFT-based segmentation and analysis of TEM images, a total of 80 reduced 2-D FFTs (2048 x 2048 pixels), collected during multiple experiments, were utilized. To standardize the dataset, the 2-D FFTs were resized through interpolation to a uniform size of 1024 x 1024 pixels. Resizing was performed in the FFT domain, rather than on the original TEM images, to preserve the full range of spatial frequencies captured in the original high-resolution images. Resizing the TEM images before FFT computation could have led to the loss of sharp features crucial for detecting fine crystallographic details. By resizing in the FFT domain, we retain the original spatial information while controlling the frequency range used for training. Furthermore, the ground truth labels for the 2-D FFTs were generated at the full resolution of 2048 x 2048 pixels and then resized to 1024 x 1024 pixels to match the input images. This ensures that no information is lost in the labels during the resizing process, maintaining the accuracy and precision of the training data. This approach also reduces computational demands and memory requirements, which were essential given the limitations of our hardware.

To increase the diversity of the dataset, the 80 2-D FFTs were augmented by applying affine transformations and image rotations, resulting in a total of 1986 images. The affine transformations used on the images include : Rotation - Randomly rotated slightly by setting a range of ±0.2 degrees, Horizontal Translation - Randomly shifted by up to 5% of the image's width in either the left or right direction, Shearing - Slanting the shape of an image, towards the left or right direction by 0.05 degrees, and Zooming - Randomly zoomed in or out by up to 5%. Data augmentation was applied to the 1024 x 1024 pixel images, which were then cropped to 1024 x 512 pixels to create our primary training dataset (**Fig. S1**). To investigate the impact of both dataset size and image sizes on model accuracy, a similar data augmentation technique was applied to generate four distinct datasets. Two datasets consisted of 1986 images each, one with a size of 256 x 256 pixels (resized through interpolation cv2.INTER_AREA[29]) and the other with 512 x 512 pixels (resized through interpolation cv2.INTER_AREA). Additionally, two smaller datasets of 500 images each were generated with the same respective image sizes using data augmentation. While all these datasets were used for model evaluation as discussed in the Results section, our final optimized model was trained using the 1024 x 512 pixel dataset. This approach ensures that the deep learning model is trained on a diverse range of images and FFT-based segmentation is more accurate.

**Programming and training machine learning models:**

The study employed Python for all programming activities, with the Keras framework of Tensorflow library [30] utilized for machine learning. The final dataset consisted of 1986 images with dimensions of 1024 × 512 pixels, randomly divided into training (90%, or 1771 images) and validation (10%, or 197 images) sets. While we utilized the reduced size images (256x256 and 512x512 pixels) for comparative analysis as shown in **Fig. S7**, these were not used for our final proposed model as they resulted in lower performance metrics (**Table S1**). The UNet-type architecture used in our study comprised four levels in both contracting and expansive paths (with the fifth level serving as the bottleneck), with each level containing two convolutional layers followed by either max-pooling (in the contracting path) or up-sampling (in the expansive path) (**Fig. S6**). The models were trained for 100 epochs on Dell workstations with Nvidia RTX A4000, 16GB, 4DP GPUs, with each model taking approximately 10-11 hours to train. The training period was deemed sufficient for experimentation with network architecture, data preprocessing, and hyper-parameter tuning. The stability of the models was ensured by tracking loss as a function of epoch number, indicating general convergence. Finally, the binary segmentation map, which classified individual pixels as particle or background, was obtained by thresholding predicted softmax output for each pixel.

**Hyperparameter Tuning for U-Net model**

Hyper parameter tuning was carried out using Bayesian Optimization via the Keras Tuner Python Library. Bayesian Optimization uses probabilistic models to guide the hyper parameter search process. It is particularly useful for black box functions and problems where the objective function takes a long time to evaluate.

Binary Cross-Entropy (BCE) function was employed as the loss function which is effective in handling imbalanced class distribution as is our case, where the region of FFT features are very small compared to the background; The formula for BCE is given by:

$$BCE\ Loss = -[y * \log(p) + (1-y) * \log(1-p)] \quad \#(6)$$

Where, $y$ is the true label (0 or 1), $p$ is the predicted probability of the positive class and log is natural logarithm.

Metrics used for the evaluation of neural network's segmentation performance are: the Dice coefficient, Intersection over Union (IoU), Pixel Accuracy, and Recall. To define the metrics, we use several key terms based on the ground truth set $G$ and the predicted set $P$. True Positives ($TP$) represent the intersection of $G$ and $P$, signifying pixels correctly identified as belonging to the target object. Conversely, False Positives ($FP$) denote pixels incorrectly predicted as positive, meaning they are present in $P$ but absent in $G$. True Negatives ($TN$) indicate pixels correctly identified as not belonging to the object, while False Negatives ($FN$) represent pixels that were actually part of the object in $G$ but were missed by the prediction and therefore not included in $P$.

The metrics are then formulated as follows:

$$Dice = \frac{2 * |TP|}{(|TP| + |FP| + |FN|)} \#(7)$$

Where | | denotes the cardinality of the set.

$$IoU = \frac{|TP|}{(|TP| + |FP| + |FN|)} \#(8)$$

$$Pixel \frac{Accuracy}{Accuracy} = \frac{|TP| + |TN|}{(|TP| + |TN| + |FP| + |FN|)} \#(9)$$

$$Recall = \frac{|TP|}{(|TP| + |FN|)} \#(10)$$

In this work, the objective for the optimization was minimizing the validation loss and 15 trials were allowed with 100 epochs per trial. A callback was added to the Tuner to ensure that if the model's validation loss did not continue to increase after 3 epochs, the training is terminated to save experimentation time.

In each trial, the optimization occurred over 6 parameters. These include the learning rate, convolutional kernel size, number of convolutional filters, convolutional transpose kernel size, activation function and 3 drop out values treated separately. The learning rate was set to vary between three values which were 1e-4, 1e-5 and 1e-6. The convolutional kernel size was set to vary between 3 and 5. The convolutional transpose kernel size was set to vary between 2 and 3. The number of convolutional filters in the first layer was set to vary between 32 and 128, with subsequent layers doubling this number as per the standard U-Net architecture. In our final implementation, we used 128 filters in the first level of the encoder, doubling at each level to reach 2048 filters at the bottleneck level. The activation function was set to vary between ReLu, eLu and GeLu. All 3 dropout values were set to vary between 0 and 0.5 with steps of 0.1.

## ■ RESULTS AND DISCUSSIONS

### FFT feature detection using deep learning – high resolution image segmentation

Deep learning technique was utilized to extract periodic features from the 2-D FFTs, effectively reducing the impact of non-periodic noise contributions in the diffraction-like graph obtained by radial profiling. Semantic segmentation is an essential technique employed in machine learning for the identification and classification of individual pixels in an image into distinct classes or entities [31]. By mimicking the way in which humans perceive and analyze visual data, semantic segmentation is a fundamental step in enabling machines to comprehend and interpret micrography images accurately. In the context of micrography, semantic segmentation holds significant potential in distinguishing between different cell types or structures at the pixel level, thereby aiding in various fields, including medical diagnosis, biological research, and particle sizing [32].

U-Net architecture is a widely used deep learning technique for semantic segmentation. U-Net is composed of two main pathways, namely, the contracting and expansive paths. The contracting path focuses on capturing the context of the input image, whereas the expansive path enables precise localization by up sampling the feature maps [33]. The U-Net segmentation model is employed to detect features in the 2-D FFT (**Fig. S6**) where very few annotated images are used. A large set of training data and corresponding ground truth are generated using data augmentation as detailed in previous sections. To optimize the training set conditions and model parameters, four factors are taken into consideration, namely, the size of the training set, size of the training image, the number of filters, and the threshold of the segmented image.

To assess the influence of training dataset size and image size on segmentation accuracy, a series of experiments were conducted. The model was trained using varying quantities of images and at multiple sizes, specifically comparing performance across two distinct dataset sizes and three image sizes.

A clear correlation between the number of training images and the model's performance was revealed. As depicted in **Fig. S7**, a noticeable improvement in the precision and clarity of feature detection was observed when the dataset size was increased [34]. These findings are further supported by the quantitative metrics presented in **Table S1**.

Furthermore, the impact of image size was examined by maintaining a fixed dataset size and varying the pixel dimensions of the training images. The results, again illustrated in **Fig. S7** and quantified in **Table S1**, demonstrated that finer details were facilitated and the issue of missing features observed at smaller image sizes was mitigated when larger image sizes were used. Notably, a consistent enhancement in feature identification was shown with the progression from the smallest to the largest image size. While improved segmentation accuracy is offered by larger

In semantic segmentation, convolutional layers with a higher number of filters can extract more complex patterns from image data effectively [35]. Therefore, this study evaluates the effect of increasing filters on the semantic segmentation task. Based on our assessment, it was determined that 128 filters in the first layer yielded better overall results for segmentation (**Fig. S8**)(**Table S2**).

Since a sigmoid activation function is used in the output layer of the U-Net model for binary segmentation, probability maps with values between 0 and 1 for each pixel are produced. A threshold must be applied to convert these probabilities into binary segmentation masks (where pixel values > threshold are classified as foreground, and values ≤ threshold as background). After the U-Net model trained on 1024 x 512 pixels images was finalized, an optimal threshold for converting the model's sigmoid output probabilities into binary segmentation masks was determined. Threshold values ranging from 0.0 to 1.0 were systematically evaluated using multiple segmentation quality metrics and 0.90 was selected as the optimal threshold (**Fig. S9-S10**).

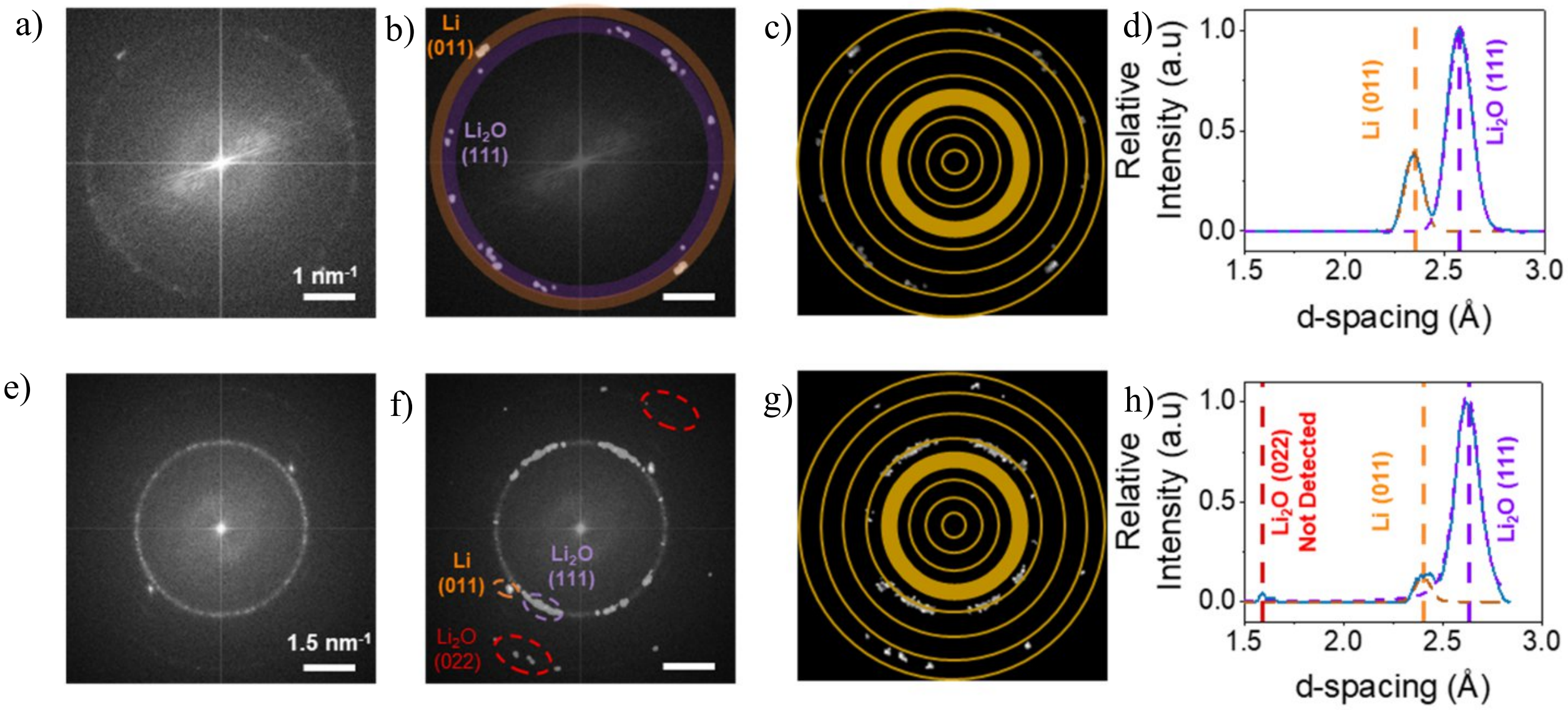

**Figure 2.** Image analysis of (a) computer generated FFT of TEM image of the SEI in a cycled Li metal anode: (b) detected FFT features using U-Net model trained with full (1024 x 1024 pixels)

images and (c) circular integration technique on segmented image to obtain the (d) diffraction-like graph of the 2-D FFT. Image analysis of (e) another computer generated FFT of TEM image of the SEI in a cycled Li metal anode: (f) model trained with full (1024 x 1024 pixels) images suffer from improper symmetry in segmentation (marked in dashed red) and merged features (marked in dashed purple). (g) circular integration technique on segmented image (h) failed peak detection in diffraction-like graph for low intensity features.

The U-Net model, as described previously, was utilized to detect features in the 2-D FFT (**Fig. 2(a)**) and generate a corresponding labeled image with white markings indicating the detected features. The labeled 2-D FFT was subsequently masked onto the original 2-D FFT (**Fig. 2(b)**), and circular integration (**Fig. 2(c)**) was applied to generate the diffraction-like graph (**Fig. 2(d)**). An example of this process is illustrated in **Fig. 2**, where the 2-D FFT obtained from a TEM image of a Li metal anode was processed to obtain the diffraction-like graph. The Li (011) (2.42Å) and $Li_2O$ (111) (2.65Å) components were detected using the peak positions and the user-defined database created using ICDD database [36]. The segmentation model, U-Net, was trained using 1024 x 1024 pixels images. However, the model suffered from poor symmetry in some cases of the resultant FFT (**Fig. 2(e)**), as demonstrated in **Fig. 2(f)** with regions marked in red dots, and imprecise segmentation due to the detected features being merged, as shown in regions marked in blue dots. Moreover, the peak detection of crystalline components using circular integration (**Fig. 2(g))** sometimes failed at very low intensity, as shown in **Fig. 2(h)**, where the $Li_2O$ (022) component was not detected due to the low intensity of the feature. **Fig. 2(h)** shows a case where peak detection failed for low-intensity features, particularly for the $Li_2O$ (022) component. This failure is primarily due to the missing detected features on the symmetrical side of the $Li_2O$ (022) component (marked in red). Since the radial profiling is done for set of pixels at a fixed radial distance from the center, the missing features affect the total intensity of the peak obtained.

**FFT half image training technique**

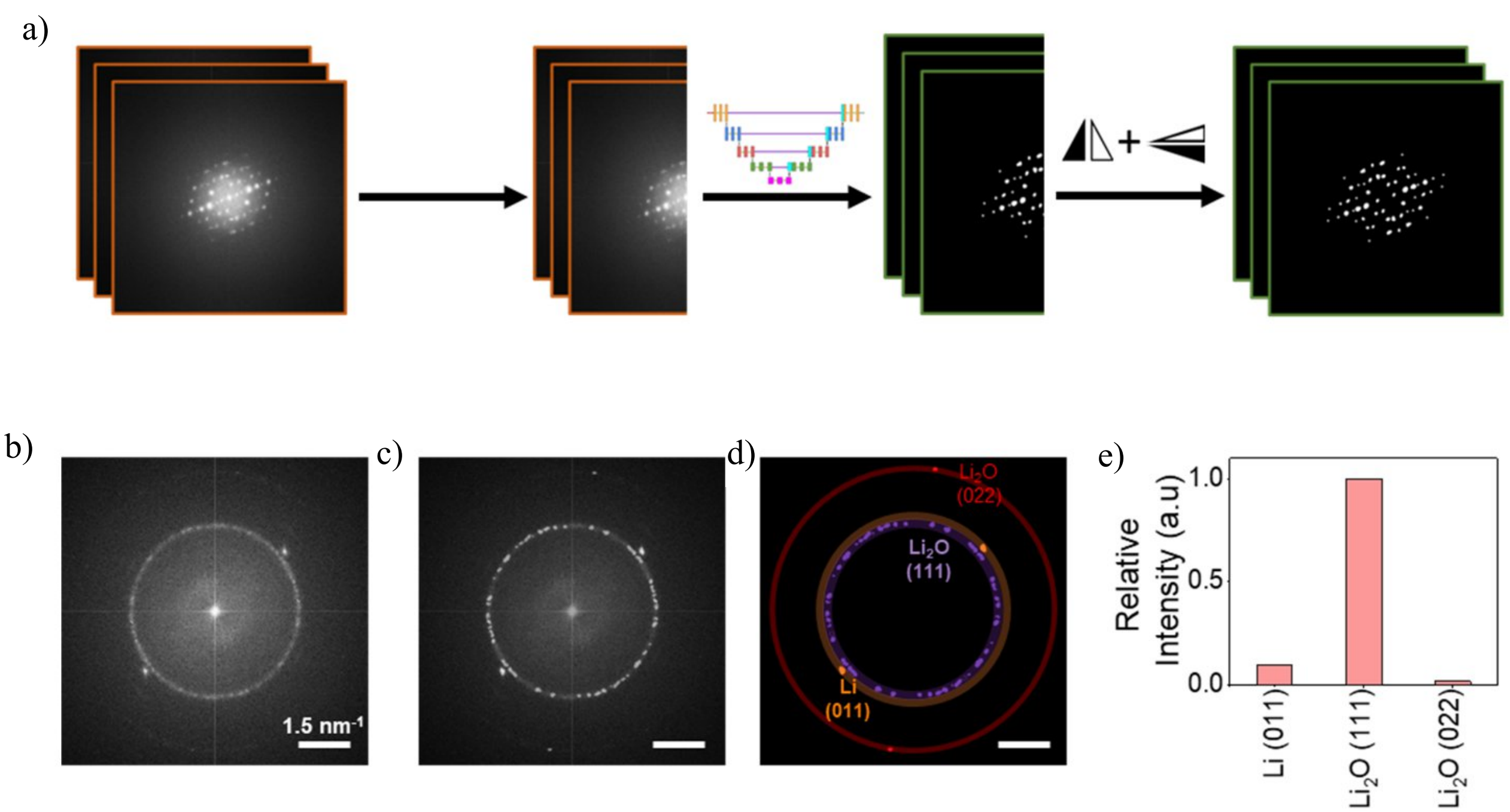

**Figure 3.** (a) Schematic of the usage of half 2-D FFTs for feature detection. Image analysis of (b) computer generated FFT of TEM image of the SEI in a cycled Li metal anode. (c) Model trained with half (1024 x 512 pixels) images preserves symmetry and produces clear distinct features. (d) Individual features detected from the segmented image with mask for IFFT. (e) Feature intensity graph generated from extracted feature properties.

| Components | d-spacing (Å) | d-spacing (Å) (from | % Match of program calculated d- | Feature size | Pixel value count |
| --- | --- | --- | --- | --- | --- |

| | (Calculated) | GATAN) | spacing with GATAN calculated | (pixels)<br>(IFFT mask) | (a.u.) |
|---|---|---|---|---|---|
| Li (011) | 2.41641 | 2.416 | 100% | 20 | 28078 |
| $Li_2O$ (111) | 2.62746 | 2.6528 | 99.04% | 17 | 287173 |
| $Li_2O$ (022) | 1.59336 | 1.593 | 100% | 11 | 4601 |

**Table I**. Tabulated results from the instance segmented image.

To address the imprecise segmentation, the symmetry of the 2-D FFT was utilized by dividing the image into two halves, using only half of the image for training, and reducing the training image size by half to 1024 x 512 pixels. The mathematical basis for this lies in the properties of the DFT for real-valued signals. From equation [1], if f(x,y) is a real-valued signal, then its DFT F(u,v) satisfies the condition $F(-u,-v) = F^*(u,v)$, where $F^*(u,v)$ is the complex conjugate of F(u,v).

The DFT of real-valued signals exhibits Hermitian symmetry, meaning that the negative frequency components are the complex conjugate of the positive frequency components. Therefore, one half of the FFT domain contains all the unique information present in the full FFT [37]. By using only half the domain, we effectively eliminate redundant information without losing any essential data. Therefore, U-Net's performance would not be significantly affected by choosing the opposite half. Using half of the FFT reduces the input size to the U-Net by 50% which not only reduces the computational load but increases the accuracy of the model (**Table S3**).

To obtain a complete 2-D FFT, half of the 2-D FFT is first duplicated and then horizontally mirrored. The mirrored image is then mirrored vertically (**Fig. 3(a)**) (**Fig. S11**). Flipping operation is carried out using cv2.flip function [38]. The challenge in processing the 2-D FFT (**Fig. 3(b)**) using model trained with full 2-D FFTs (1024x1024 pixels) was overcome by employing a model trained on half 2-D FFTs (1024x512 pixels) for feature detection, as shown in **Fig. 3(c)**. Following the initial segmentation process, individual regions were identified and characterized using connected component labeling. The cv2.connectedComponents[39] function (OpenCV library) was applied to the resulting binary image to assign unique labels to each contiguous region (**Fig. S14(b)**) (**Fig. 3(d)**). Subsequently, region properties, including centroid, major axis length, and minor axis length, were extracted using the skimage.measure.regionprops_table[40] function (scikit-image library), which were then used to obtain the intensity profile after masking (**Fig. 3(e)**). The centroid was used as the center of each extracted feature, and the size of each feature was calculated as:

$$Size\ of\ the\ feature = \sqrt{\left(Major_{axis_{length}}{}^{2} + Minor_{axis_{length}}{}^{2}\right)} \# (11)$$

Although the region properties are initially calculated as floating-point values, they are converted to integers for subsequent processing.

The components detected from the instance segmentation provide not only the exact position of the components detected, which agrees closely with the values generated from the GATAN Micrograph software. This size information can be used to design masks of the appropriate size to prevent noise from the FFT from being included in the mask, and a more accurate and discrete intensity profile was obtained after masking. **Table 1** presents a comparison between our automated measurements and those obtained manually using GATAN software, a standard in the field. This comparison validated our method's accuracy, as demonstrated by a >99% d-spacing match, and confirmed its reproducibility across varied FFT generation methods. Through this validation, the reliability of our automated workflow was established, positioning it as an alternative to time-consuming manual analysis.

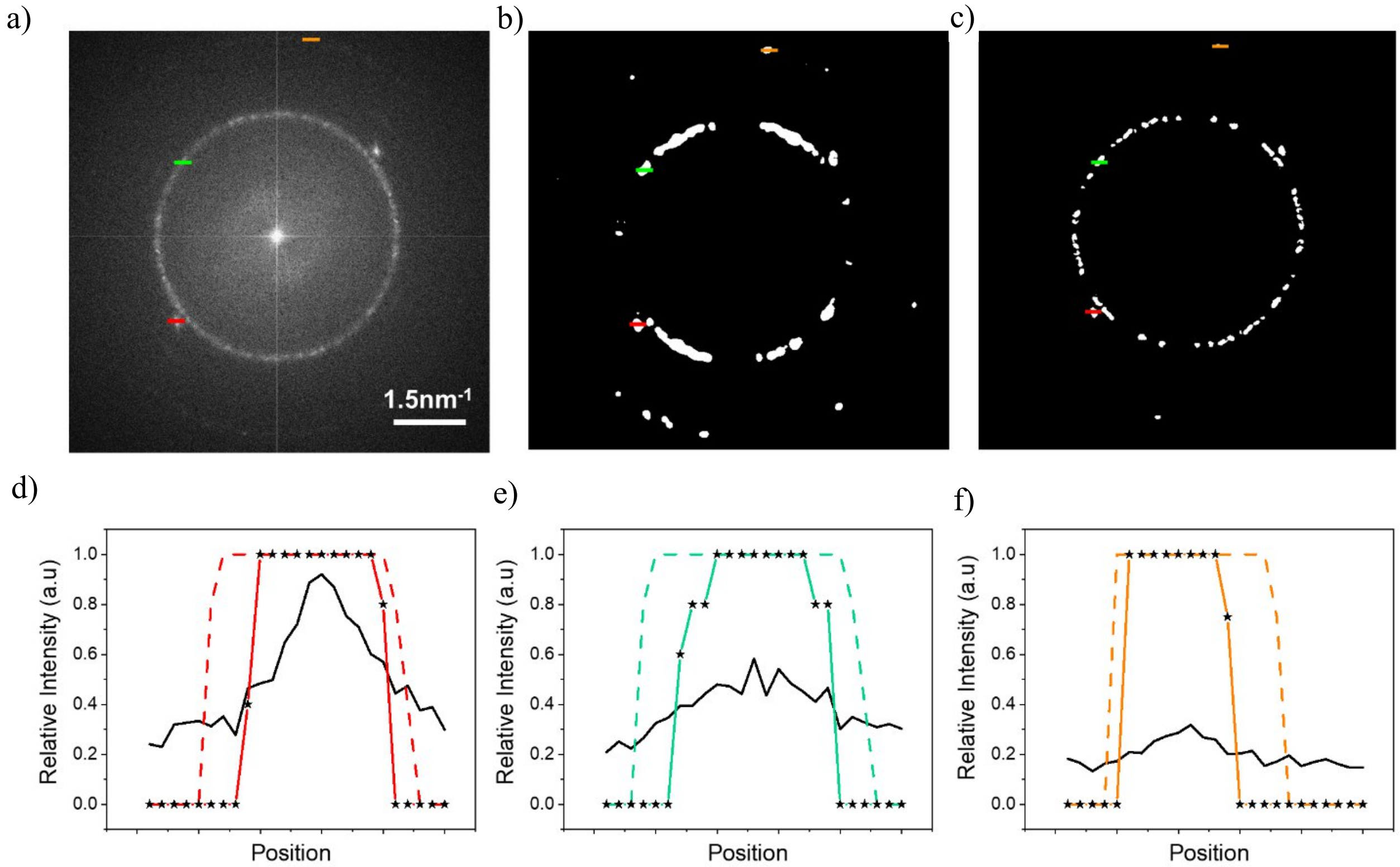

**Figure 4.** Intensity profiles for selected features in (a) computer generated FFT of TEM image of the SEI in a cycled Li metal anode. Line scans of three features compared with segmented images generated by (b) model trained with full (1024 x 1024 pixels) images and (c) model trained with half images (1024 x 512 pixels). (d-f) Intensity variation for each feature in the raw image (solid black line), model trained with full (1024 x 1024 pixels) images (dashed), and model trained with half images (1024 x 512 pixels) (starred-solid).

The features detected by the model trained with full size (1024x1024 pixels) images tend to produce features at least 30% bigger than the model trained with half-size images (**Fig. 4 (a-c)**).

The intensity profile shows the robustness of the new model trained with half images (**Fig. 4 (d-f)**). The size of the segmented feature is important as a larger sized feature makes the distinct features more prone to merging, resulting in inaccurate IFFT generation from the masking. To quantify the closeness in shape and size between the segmented feature and its ground truth, we utilize the Hausdorff distance[41]. This measure determines the maximum mismatch between two sets of points, representing the boundaries of the segmented feature (S) and its corresponding ground truth (G). The Hausdorff distance (H(S, G)) is defined as:

$$H(S,G) = max\{h(S,G), h(G,S)\} \# (12)$$

Where:

$$h(S,G) = \max_{s \in S} \min_{g \in G} ||s - g|| \# (13)$$

$$h(G,S) = \max_{g \in G} \min_{s \in S} ||g - s|| \# (14)$$

Where, $s$ is a point in the set $S$, $g$ is a point in the set $G$ and $||s - g||$ is the Euclidean distance between the points $s$ and $g$. $h(S,G)$ calculates the maximum distance from a point in the segmentation to its closest point in the ground truth. $h(G,S)$ calculates the maximum distance from a point in the ground truth to its closest point in the segmentation. The Hausdorff distance $H(S,G)$ takes the maximum of these two values, representing the worst-case mismatch between the two sets. A lower Hausdorff distance value signifies a greater conformity between the segmented feature and it's corresponding ground truth. In the ideal scenario of perfect segmentation, where the segmented features and ground truth are indistinguishable, the Hausdorff distance would be 0.

U-Net model trained with 1024 x 512 pixels 2-D FFTs was benchmarked against: (1) U-Net model trained with 1024 x 1024 pixels 2-D FFTs, (2) CNN model trained with 1024 x 512 pixels images, and (3) K-Means clustering algorithm optimized for 1024 x 512 pixels images. Evaluation metrics employed were the Dice coefficient, IoU, pixel accuracy, precision, recall, and Hausdorff distance (**Fig. S12**). Descriptions of the CNN architecture and the K-Means clustering method are provided in **Section S1** of supplementary information. Superior performance across all metrics, including the Hausdorff distance, was observed for the U-Net model trained with 1024 x 512 pixels 2-D FFTs (**Table S4**).

An ablation study was also conducted to investigate the impact of 2-D FFT pre-processing on U-Net model performance (**Fig. S13**). The performance of U-Net models trained with 2-D FFTs at various stages of pre-processing was compared. A positive correlation was observed between model performance and the similarity score of the pre-processed 2-D FFT to the GATAN-generated 2-D FFT (**Table S5**)(**Fig. S3**).

**Component detection and mapping**:

Features detected in the 2-D FFT were matched to corresponding components in the database based on their d-spacing values. The component with the nearest d-spacing value was assigned to each feature. Subsequently, features were grouped by component, and the d-spacing values and particle sizes within each group were averaged to yield a single representative value for each component (**Fig. S14(c)**). The markers corresponding to the grouped components are used to reconstruct 2-D FFTs with isolated features (**Fig. S14(d,e)**).

IFFT (Inverse FFT) was done on the masked 2-D FFTs of isolated components to map the component distribution on the TEM image. The IFFT generated had regions of varied intensity and indistinct boundaries between regions of interest (**Fig. S15**). Watershed segmentation[42] was utilized to extract bright regions of interest from IFFT maps. The segmentation was done in three steps: (1) initial binary separation was performed using Otsu's automated thresholding with binary inversion, (2) noise was removed via morphological opening operations using a square kernel, and (3) background regions were identified through dilation to ensure clear structure separation. Parameters of watershed segmentation was optimized through a systematic grid search across multiple parameter combinations, wherein kernel size (3×3, 5×5, 7×7), opening iteration count (0, 1, 2), and dilation iteration count (0, 1, 2) were varied. Each combination was evaluated against 15 manually annotated ground truth images using five quantitative metrics: Dice coefficient, IoU, pixel accuracy, precision, and recall. Parameters were selected based on the highest average rank across all metrics (**Fig. S16**). Through comprehensive evaluation, the optimal parameter configuration was found to be a 7×7 kernel size with 2 opening iterations and 0 dilation iterations. It was observed that configurations with low opening iterations (0) combined with high dilation iterations (2) produced poor segmentation results, as insufficient noise removal prior to dilation caused boundary expansion of noise artifacts, inappropriate region merging, and resulting in a completely dark image (**Fig. S17**).

Given the robust model, this tool can be used for high throughput TEM image analysis. The detection of the periodic components and the mapping of the components can be achieved with the tool (**Fig. S18**).

**LiF beam damage analysis**

The investigation of the SEI in lithium-ion batteries requires the use of TEM for atomic resolution imaging. Electron beam and environmental factors are known to cause irradiation damage in the SEI. To avoid such damage, the lithium sample must be maintained at low temperatures, but even then, the dose rate and exposure time can impact the sample products and its rate [7]. The present study aims to analyze the beam damage of lithium fluoride (LiF), which is a commonly occurring SEI component, using a high-volume TEM image processing tool. To this end, we obtained data on LiF beam damage from a study that investigated the effects of beam damage on the imaging of SEI components in a cycled lithium metal anode, while considering the dose rate and exposure time. In our study, we analyze the images of the LiF particles at cryogenic and room temperatures, and with various dose rates ranging from 50 e $A^{-2}$ $s^{-1}$ to 1000 e $A^{-2}$ $s^{-1}$ using the developed TEM processing methodology. The TEM images are obtained by the Mel-Build holder using an optimized workflow.

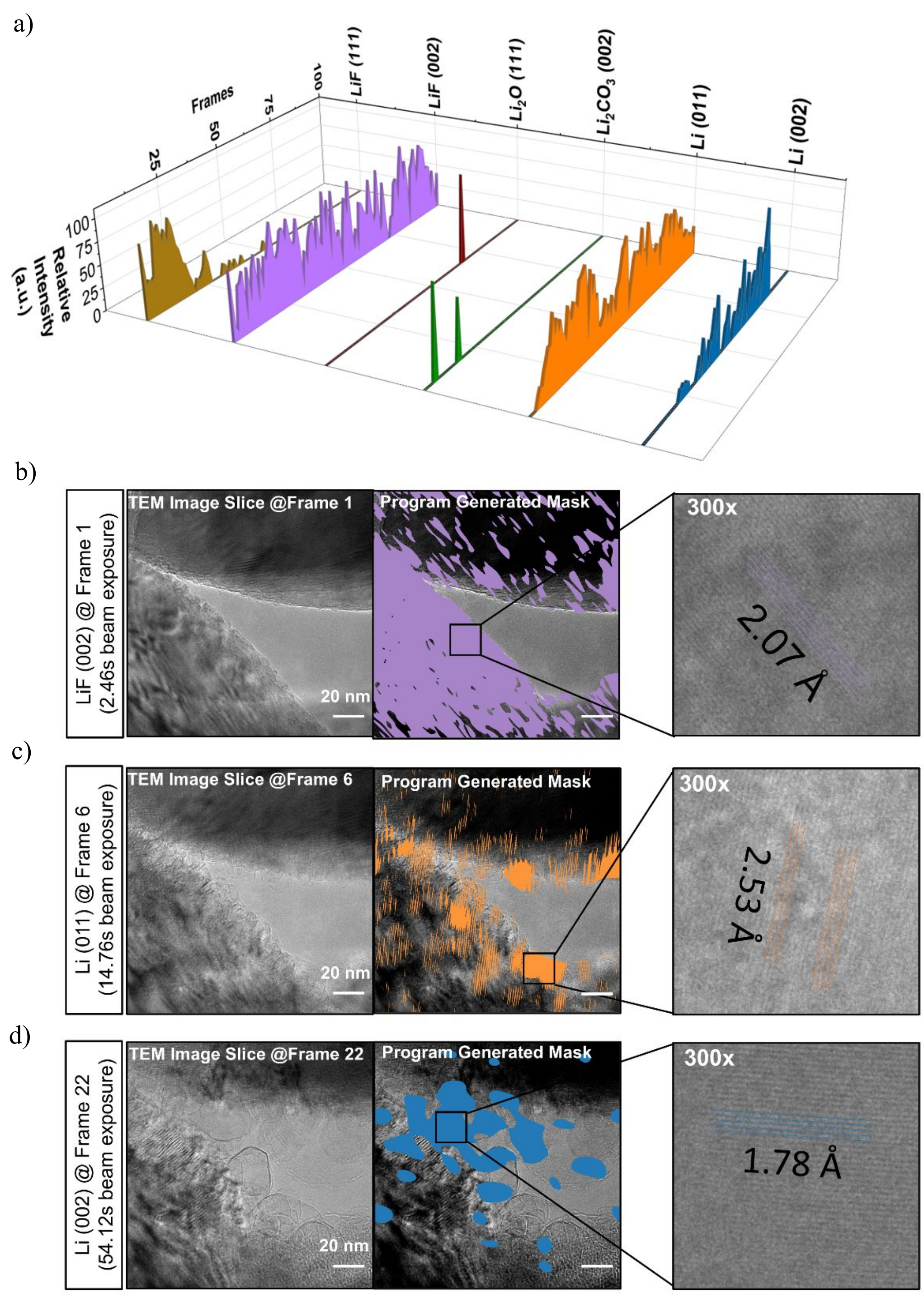

**Figure 5.** (a) Intensity distribution of detected components over 100 frames of LiF at electron dosage of 1000 e $A^{-2}$ $s^{-1}$ and cryo temperature. TEM image slice and program generated IFFT

mapping at the initial detection of (d) LiF (002) – after 2.46 s beam exposure, (b) Li (011) – after 14.76 s beam exposure, and (c) Li (002) – after 54.12 s of beam exposure.

A video was recorded during TEM imaging to monitor the beam damage of LiF particles. The Velox[43] software was used to save the resulting video file in the mrc format, which contained 100 slices corresponding to 2.46 seconds of electron beam exposure each. The mrc python library [44] was utilized to extract image data from the file, but since it was unable to retrieve the pixel size information, this information was manually entered along with the mrc file for subsequent processing. Batch file processing was employed for the automatic analysis of the components present in the sample, which were mapped using feature extraction from the 2-D FFT. The intensity of each component was recorded and compared to its corresponding intensity in the subsequent slice to track the evolution of the components during beam damage (**Fig. 5 (a)**). This allows for effective assessment of the beam damage of LiF particles during TEM imaging. The analysis of the intensity distribution of the detected components over 100 frames of beam exposure of LiF (**Fig. 5 (a)**) reveals that the Li (011) phase diffuses out of the LiF first at 6$^{th}$ frame or after 14.76 s of beam exposure (**Fig. 5 (b)**). The Li (011) facet is the most commonly observed one during nucleation, despite the slightly lower surface energy of Li (002) facet which is observed at 22$^{nd}$ frame or after 54.12 s of beam exposure (**Fig. 5 (c)**). This is due to the lower thermodynamic overpotential required to obtain Li (011) facet, which overpowers the surface energy effect [45]. The intensity of the Li (011) phase decreases and starts to increase midframe due to the appearance of underlying LiF particles. $Li_2O$ (111) and $Li_2CO_3$ (002) phases are also observed, which could be attributed to the surrounding environment of the TEM chamber [46]. LiF (002) is dominant over LiF (111) phase because it has a lower surface energy and a higher binding energy [47]. LiF (111) is observed at the initial frames of the beam exposure (**Fig. S19**) and fades away as Li (011) phase grows. There is no information provided in literature that directly answers whether using high dose rates during TEM can convert one phase with lower surface energy to one with higher surface energy. However, it is known that high dose rates can cause radiation damage to the sample, which can affect the crystal structure and properties of the material being studied. Therefore, it is possible that high dose rates during TEM could affect the surface energy of a material, but further research is needed to determine the exact mechanism and conditions under which this could occur. The consistent intensity of the (002) peak in LiF is observed to be a result of the presence of newly formed LiF particles on the bottom surface after the degradation of the topmost LiF layer due to beam damage. This observation is corroborated by the increase in the intensity of the Li (011) phase, which subsequently decreases but then increases again. The increase in the intensity of the Li (011) phase is indicative of the formation of Li (011) due to the degradation and resurfacing of LiF particles. The degradation of LiF into Li is observed to be very limited at low dose rate and at cryo temperature (**Fig. S19**). A Python-based GUI tool has been designed and developed for future use and broader distribution. This intuitive interface offers a potential advantage over current state-of-the-art analysis programs, particularly when dealing with large datasets. The GUI aims to streamline the analysis workflow, making the advanced capabilities of the program widely accessible to a wider range of users and facilitating efficient processing of complex TEM data (**Fig. S20**).

## ■ CONCLUSIONS

This paper introduces a novel deep learning-based framework for identifying and evaluating the phases present in TEM images of the SEI of Li metal anodes. The framework utilizes the FFT diffractograms of the TEM images to analyze the feature positions and determine the SEI components. The proposed method also incorporates techniques to handle high-resolution images and exploit the symmetry of the FFT diffractograms for better model performance. The effectiveness of the framework is further improved by introducing additional training data. Additionally, a comprehensive analysis of the TEM images through intensity profiling and component mapping provides valuable insights into the SEI component evolution during imaging.

In future studies, we propose utilizing the diffraction pattern generated from high frame rate TEM image patches for more detailed and accurate analysis. This alternative approach is superior to the sliding window technique for generating FFT diffractograms from TEM images as the latter may lack clarity and clear features. Furthermore, our proposed deep learning model can be applied to analyze not only the SEI components of Li metal anodes, but also other composite systems involving periodic components (Na metal anodes, Si anodes, etc.) provided an appropriate database with the respective compounds is used. The developed workflow can be further improved by integrating it for operando detection and mapping of the phases, thereby enabling the analysis of component evolution during TEM imaging.

## ■ SUPPLEMENTARY INFORMATION

The Supplementary Information is available free of charge.

The Supplementary Information contains details of the optimization of the model.

## ■ AUTHOR INFORMATION

**Corresponding Author**

*(MZ) E-Mail: miz016@eng.ucsd.edu

*(YSM.) E-mail: shirleymeng@uchicago.edu

**ORCID:**

Minghao Zhang: 0009-0002-8303-3942

Ying Shirley Meng: 0000-0001-8936-8845

## ■ ACKNOWLEDGEMENTS

This work is supported by the funding and collaboration agreement between UCSD and Thermo Fisher Scientific on Advanced Characterization of Energy Materials. TEM was performed at the San Diego Nanotechnology Infrastructure (SDNI), a member of the National Nanotechnology Coordinated Infrastructure, which is supported by the National Science Foundation (grant ECCS-1542148). The workstation to build the model was provided by the University of Chicago under YSM's startup package. F.A. acknowledges support from the Research Experience for Nigerian Engineering Undergraduates (RENEU) program at the University of Chicago.

## ■ AUTHOR CONTRIBUTIONS

G.R., M.Z., and Y.S.M. conceived the ideas. B.H. and S.B. performed TEM imaging. G.R. and F.A developed the model for TEM processing. G.R, F.A and W.W worked on the development of GUI application. All authors discussed the results and commented on the manuscript.

## ■ COMPETING INTERESTS

All authors declare no financial or non-financial competing interests.

## ■ DATA AVAILABILITY

The datasets generated and/or analysed during the current study are available in the *TEMGUILESC* Github repository (https://github.com/ganyguru/TEMGUILESC).

## ■ CODE AVAILABILITY

The underlying code and training/validation datasets for this study is available in *TEMGUILESC* Github and can be accessed via this link (https://github.com/ganyguru/TEMGUILESC).

## SUPPLEMENTARY INFORMATION

# Deep learning assisted high resolution microscopy image processing for phase segmentation in functional composite materials

*Ganesh Raghavendran[a], Bing Han[a], Fortune Adekogbe[d], Shuang Bai[a], Bingyu Lu[a], William Wu[e], Minghao Zhang[c,*], Ying Shirley Meng[a,c,*]*

a Department of NanoEngineering, University of California San Diego, La Jolla, CA 92093

b Materials Science and Engineering, University of California San Diego, La Jolla, CA 92093

c Pritzker School of Molecular Engineering, University of Chicago, Chicago, IL 60637

d Department of Chemical and Petroleum Engineering, University of Lagos, Lagos, Nigeria 101017

e Department of Computer Science, University of California San Diego, La Jolla, CA 92093

Correspondence to miz016@uchicago.edu; shirleymeng@uchicago.edu

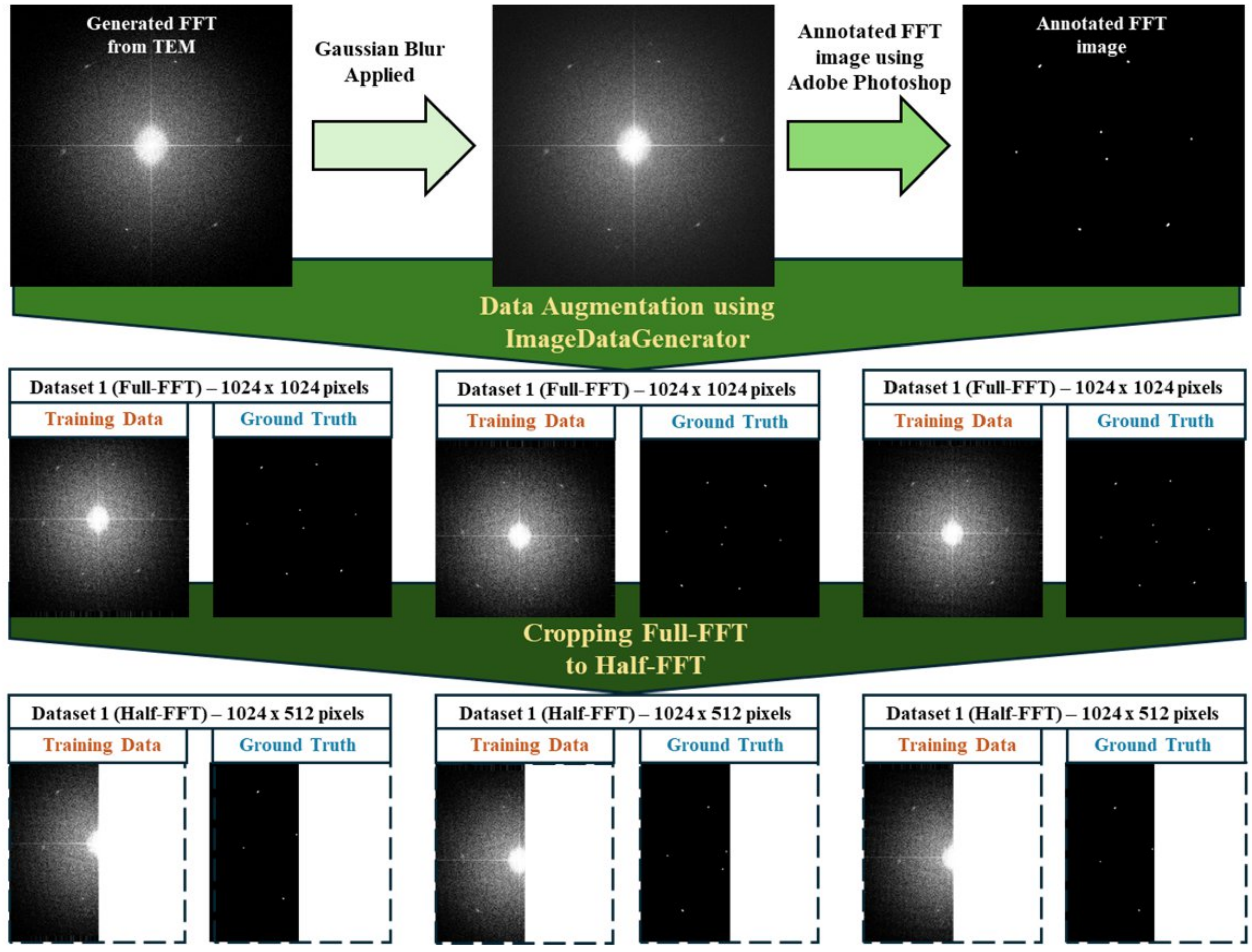

**Figure S1.** Schematic of the data generation for training the model.

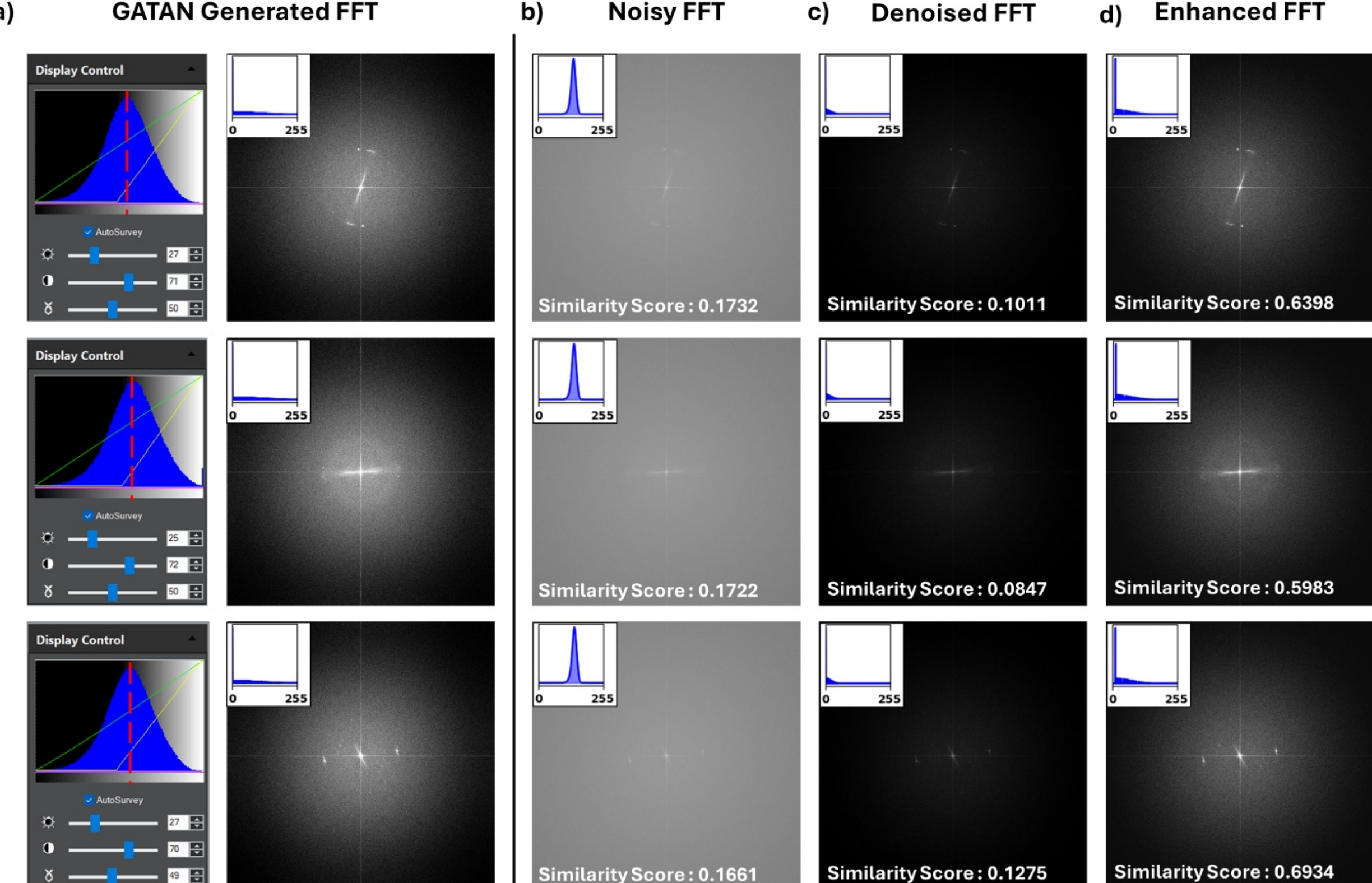

**Figure S2.** Comparison of FFT processing methods: (a) GATAN software output with display controls, (b) noisy Python-generated FFT, (c) denoised FFT, and (d) enhanced FFT. Inset histograms and similarity scores are shown for each image.

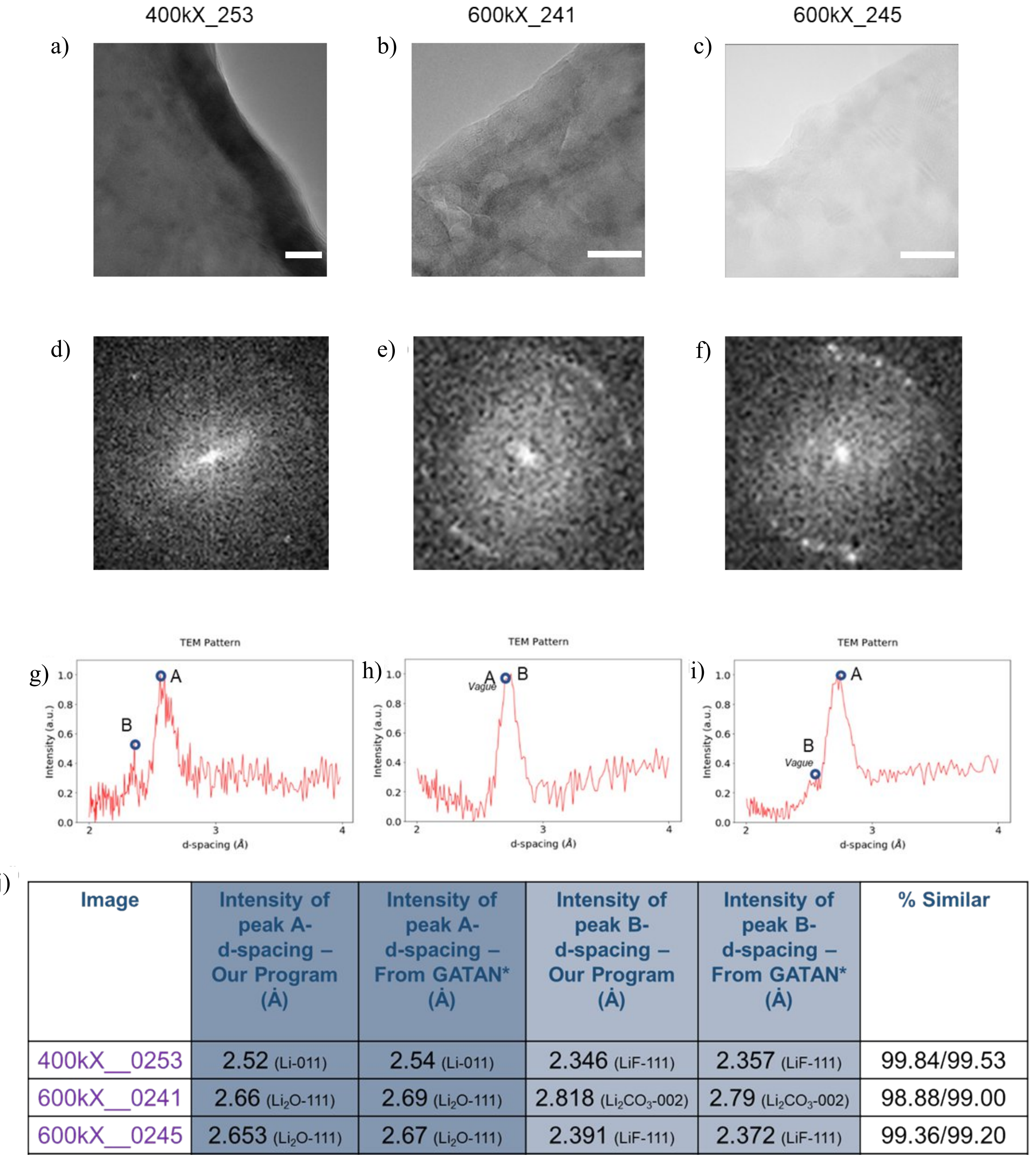

| Image | Intensity of peak A-d-spacing – Our Program (Å) | Intensity of peak A-d-spacing – From GATAN* (Å) | Intensity of peak B-d-spacing – Our Program (Å) | Intensity of peak B-d-spacing – From GATAN* (Å) | % Similar |
|---|---|---|---|---|---|
| 400kX__0253 | 2.52 (Li-011) | 2.54 (Li-011) | 2.346 (LiF-111) | 2.357 (LiF-111) | 99.84/99.53 |
| 600kX__0241 | 2.66 ($Li_2O$-111) | 2.69 ($Li_2O$-111) | 2.818 ($Li_2CO_3$-002) | 2.79 ($Li_2CO_3$-002) | 98.88/99.00 |
| 600kX__0245 | 2.653 ($Li_2O$-111) | 2.67 ($Li_2O$-111) | 2.391 (LiF-111) | 2.372 (LiF-111) | 99.36/99.20 |

**Figure S3.** (a-c) TEM images (scale bar: 10 nm) of SEI of cycled Li metal anodes. (d-f) FFT images generated for corresponding TEM images. (g-i) Diffraction-like graph generated using circular reintegration method. (j) Tabulated results obtained by manually identifying the peaks and compared with d-spacing obtained from GATAN software.

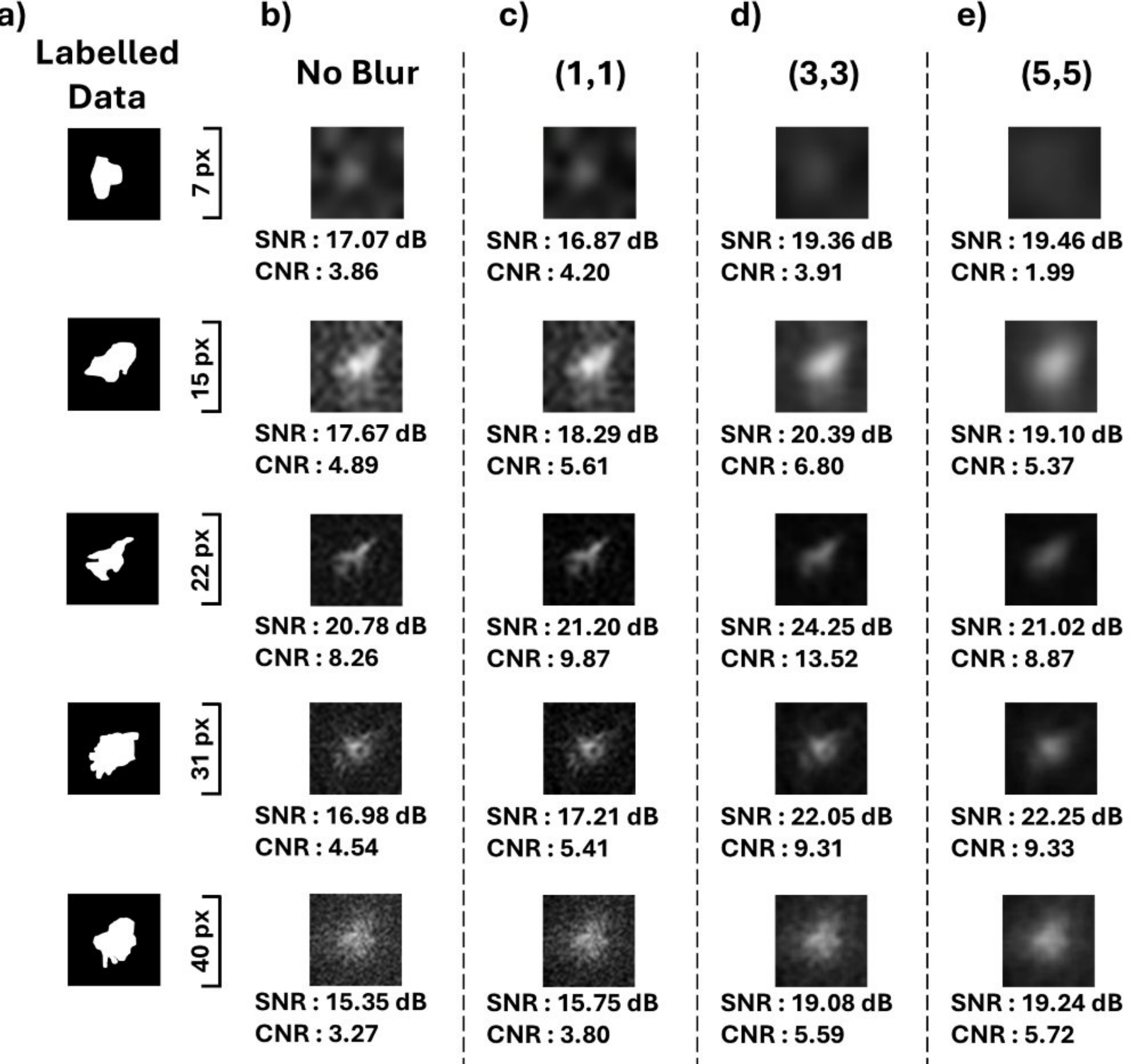

**Figure S4.** Effect of Gaussian filtering with different kernel sizes on FFT features of various sizes. **(a)** Labelled data. **(b)** No blur. **(c)** (1,1) kernel. **(d)** (3,3) kernel. **(e)** (5,5) kernel. SNR and CNR values are shown below each image.

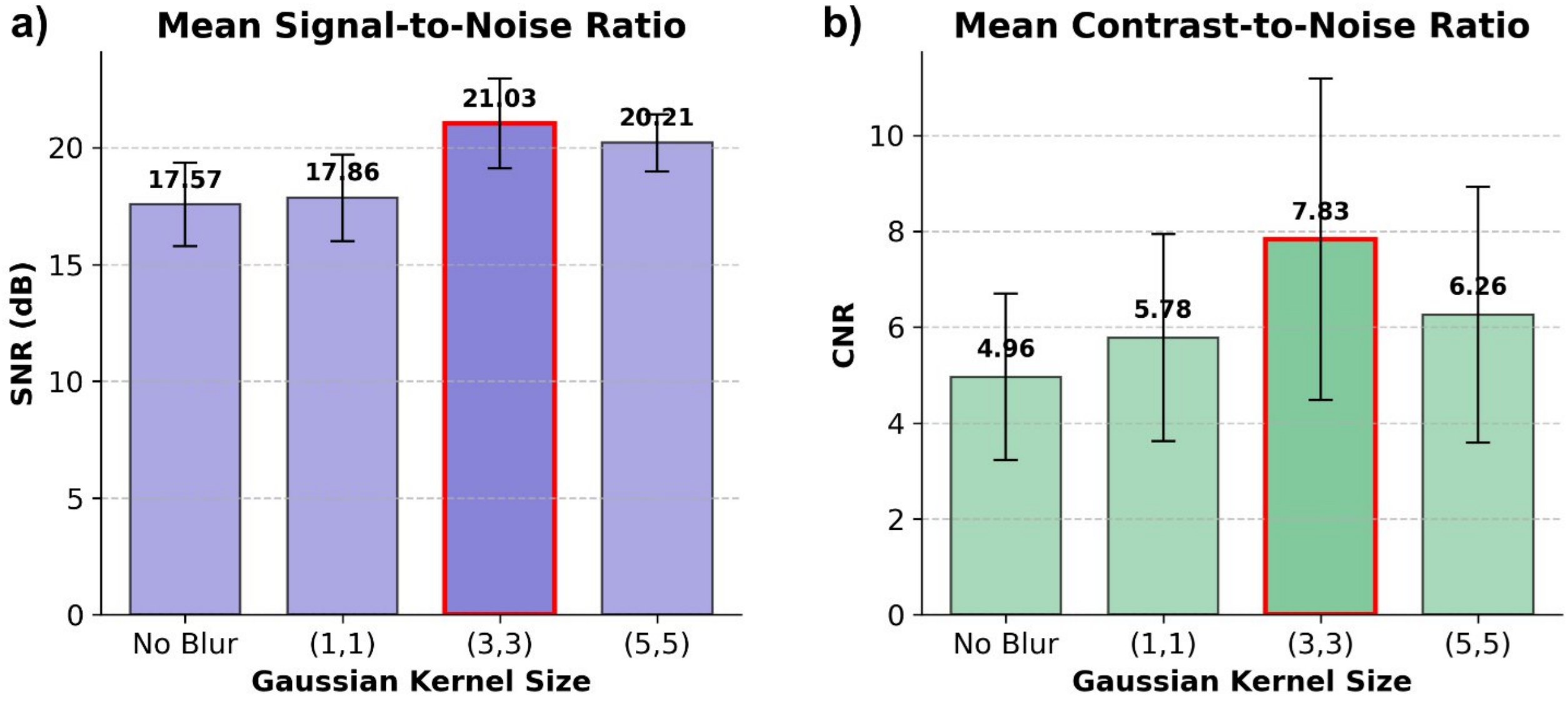

**Figure S5.** Quantitative analysis of Gaussian blur filtering with different kernel sizes on HRTEM images. **(a)** Mean Signal-to-Noise Ratio (SNR) for different kernel sizes. **(b)** Mean Contrast-to-Noise Ratio (CNR) for different kernel sizes.

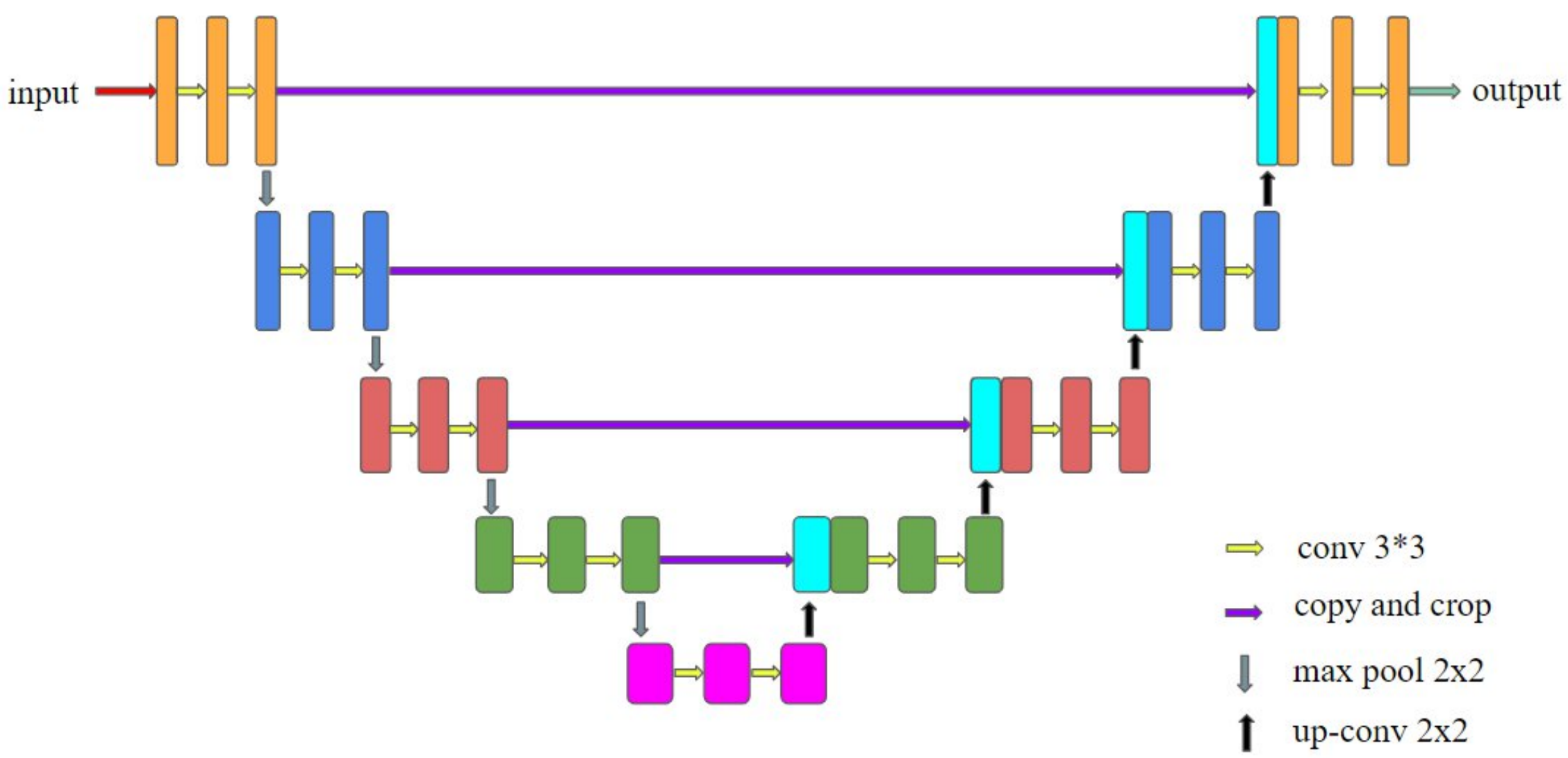

**Figure S6.** Schematic representation of the UNet-type architecture used.

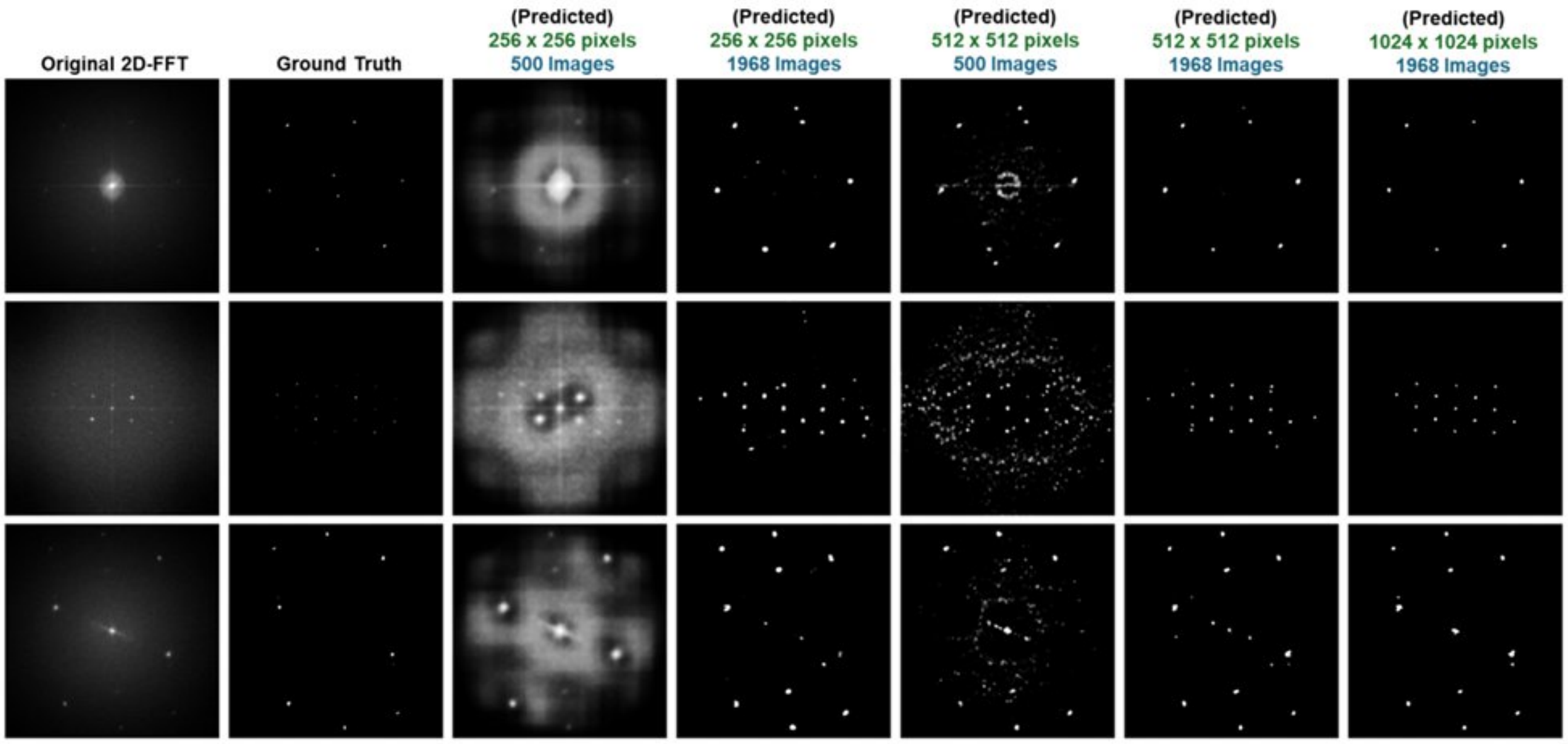

**Figure S7.** Results of the predicted FFT features using UNet model trained using various input image shape and number of input images.

| Size of the image (pixels x pixels) | No. of images | Dice Coefficient | IoU | Pixel Accuracy | Precision | Recall |
|---|---|---|---|---|---|---|
| **256 x 256** | **500** | 0.009 | 0.005 | 0.903 | 0.005 | 0.629 |
| **256 x 256** | **1968** | 0.267 | 0.165 | 0.995 | 0.163 | 0.861 |
| **512 x 512** | **500** | 0.143 | 0.078 | 0.992 | 0.080 | 0.817 |
| **512 x 512** | **1968** | 0.355 | 0.225 | 0.997 | 0.227 | 0.941 |
| **1024 x 1024** | **1968** | 0.446 | 0.297 | 0.998 | 0.304 | 0.918 |

**Table S1.** Metrics of the predicted FFT features using UNet model trained using various input image shape and number of input images.

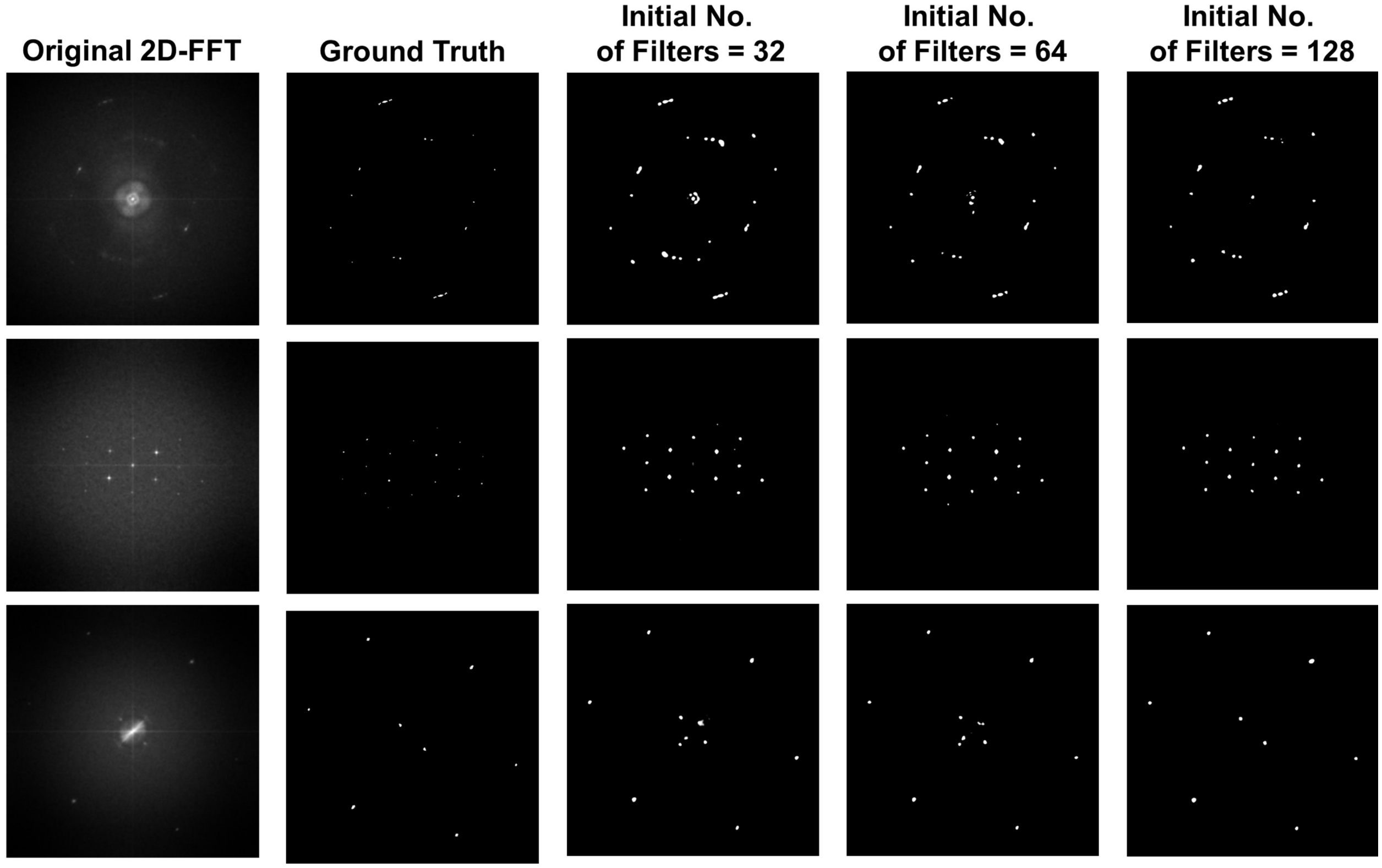

**Figure S8.** Results of the predicted FFT features using UNet model trained using various initial number of filters.

| Initial No. of Filters | Dice | IoU | Accuracy | Recall | Precision |
|---|---|---|---|---|---|
| **32 filters** | 0.396 | 0.294 | 0.998 | 0.918 | 0.293 |
| **64 filters** | 0.419 | 0.294 | 0.998 | 0.918 | 0.295 |
| **128 filters** | 0.445 | 0.296 | 0.998 | 0.918 | 0.303 |

**Figure S2.** Metrics of the predicted FFT features using UNet model trained using various initial number of filters.

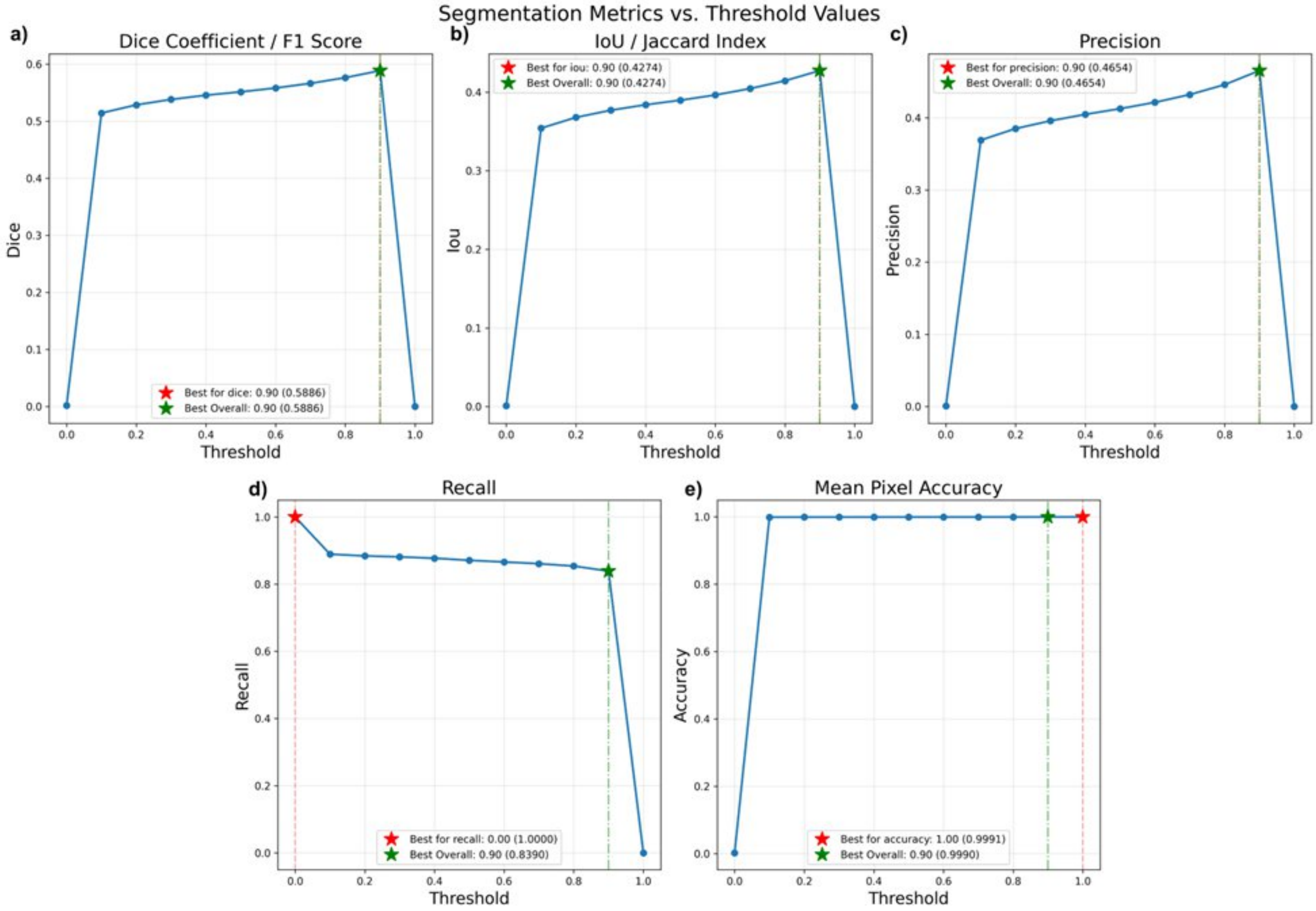

**Figure S9.** Segmentation performance vs threshold values. (a) Dice Coefficient/F1 Score, (b) IoU/Jaccard Index, (c) Precision, (d) Recall, and (e) Mean Pixel Accuracy. Red stars indicate metric-specific optimal thresholds; green stars show best overall threshold (0.9).

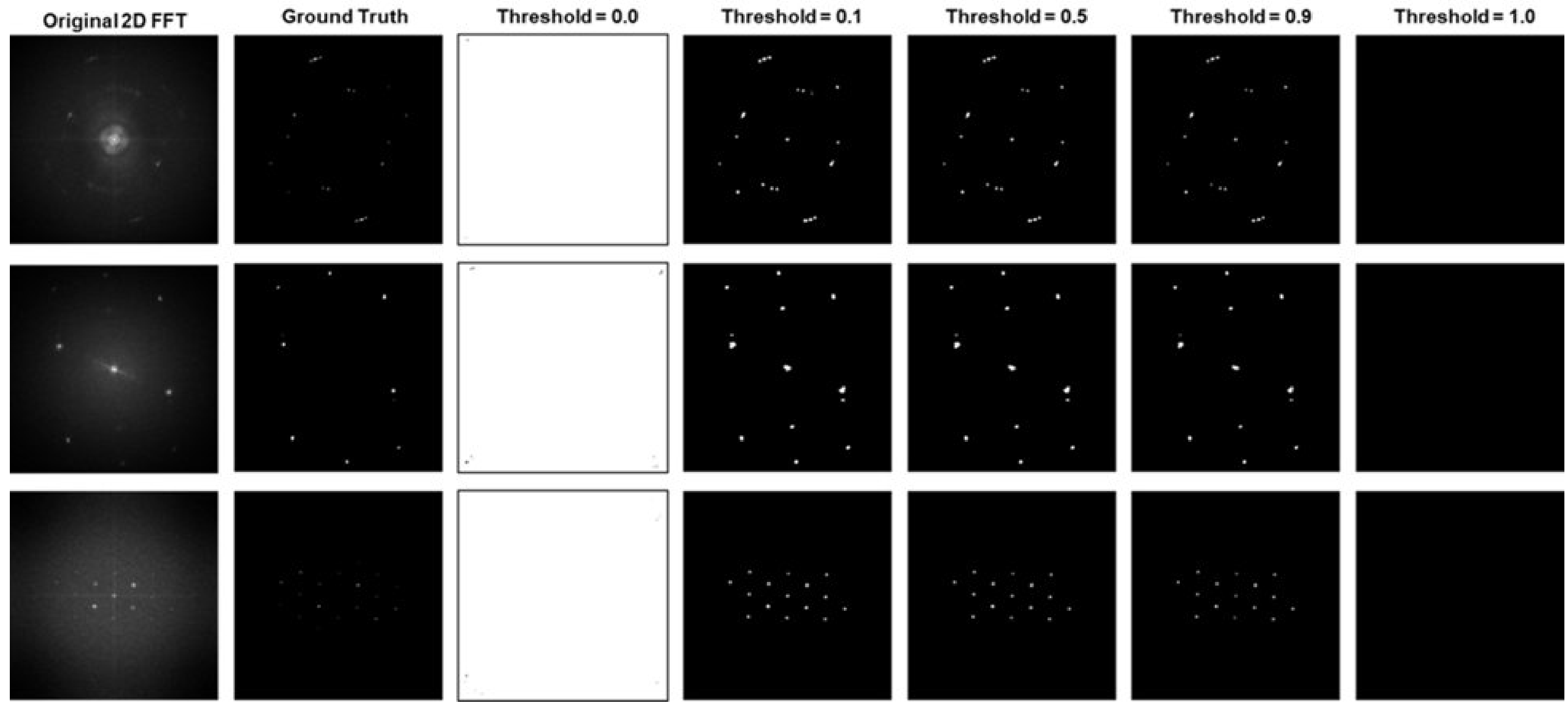

**Figure S10.** Threshold value selection for FFT feature segmentation : Original FFT images, Corresponding ground truth, segmentation at threshold 0.0, 0.1, 0.5, 0.9, and 1.0.

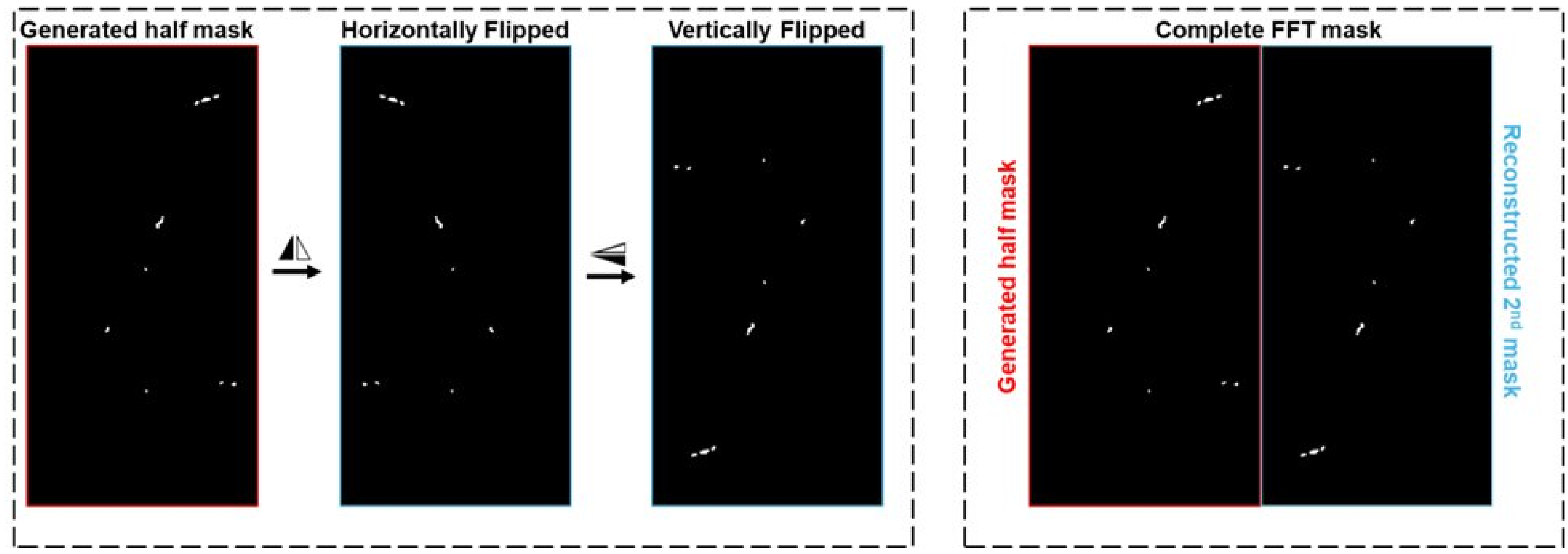

**Figure S11.** Reconstruction workflow of half-FFT image to full-FFT image

| Image Size (pixels) | No. of images used for training | Inference memory - GPU (MB) | Inference time (s) |
|---|---|---|---|
| 256x256 | 500 | 0.5 | 3.09 |
| 256x256 | 1986 | 0.5 | 3.28 |
| 512x512 | 500 | 2 | 4.82 |
| 512x512 | 1986 | 2 | 4.86 |
| 1024x1024 | 1986 | 8 | 11.48 |
| 1024x512 | 1986 | 4 | 7 |

**Table S3.** Comparison of model inference memory and inference time of 1 input data.

## Section S1 : Description of K-means clustering and CNN model used

### K-Means Clustering:

For optimizing the K-means segmentation parameters, a grid search was conducted using 50 sets of Fast Fourier Transform (FFT) images and their corresponding masks. The search explored the following parameter ranges: the number of clusters (k) [2, 3, 4, 5], morphological closing kernel sizes [(3, 3), (5, 5), (7, 7)], closing iterations [1, 2, 3], and target cluster selection methods ('brightest', 'darkest', 'auto'). The Dice coefficient was used as the evaluation metric. The optimization process yielded the following best parameters: k=5, morphological kernel size (3, 3), 1 closing iteration, and 'auto' for target cluster selection, resulting in a Dice score of 0.0353.

### CNN (Convolution Neural Network):

The proposed CNN architecture (without skip connections) maintains the same encoder-decoder structure as the traditional U-Net model while making one significant modification. Both architectures utilize an identical contracting path with five sequential levels, each consisting of dual 3×3 convolutional layers with ReLU activation, progressive dropout rates (0.1-0.3), followed by 2×2 max pooling operations that halve spatial dimensions while increasing feature channels (maximum number of filters as 2048 at bottleneck layer). Similarly, both models employ an expansive path with up-convolutions that restore spatial dimensions. However, the critical difference lies in the absence of skip connections. The model is compiled using the Adam optimizer with a learning rate of 0.0001, binary cross-entropy loss, aligning with the UNet training setup. 1968 FFT images of size 1024 x 512 pixels were used for training (Same as UNet).

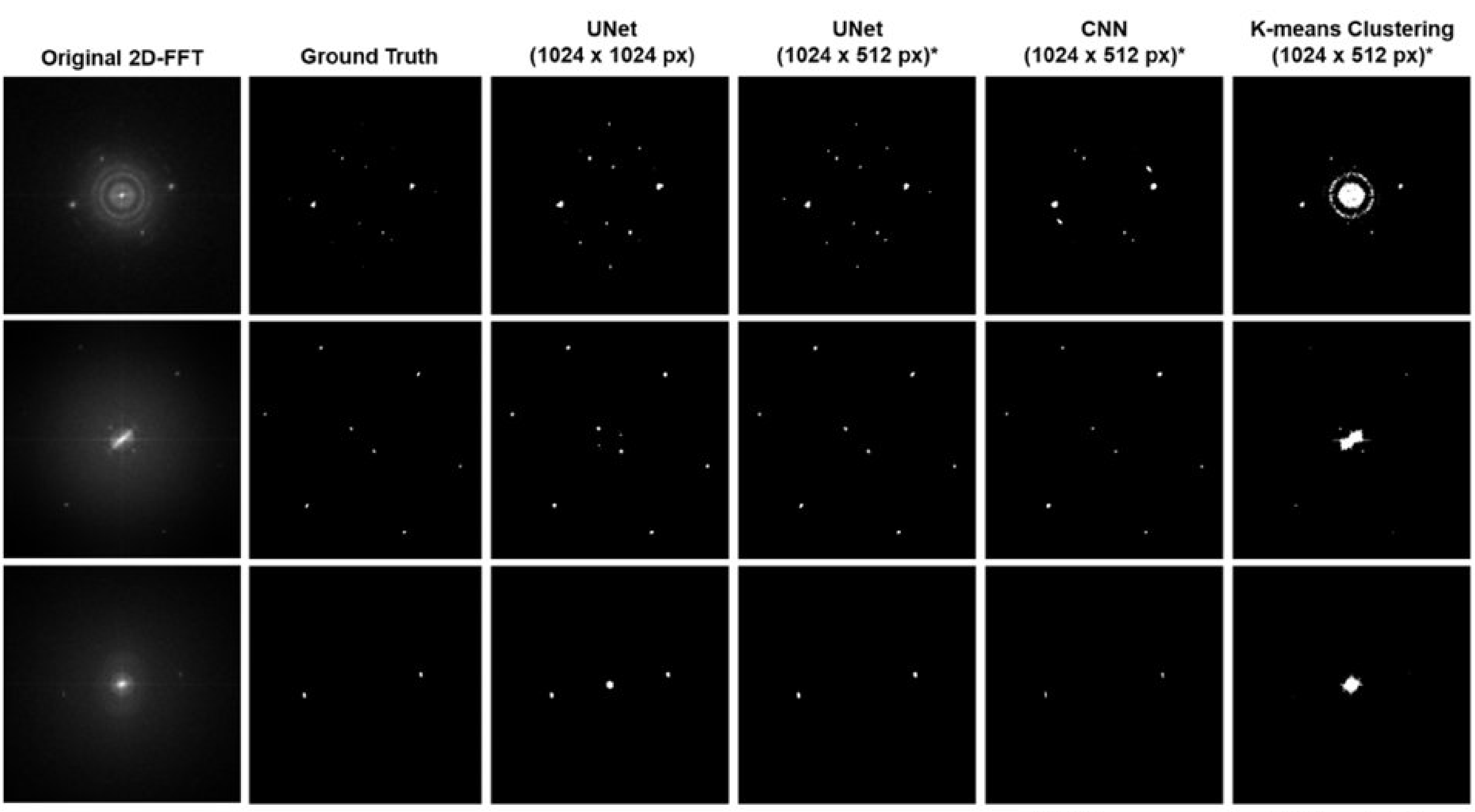

**Figure S12.** Segmentation results demonstrating the performance of U-Net (1024x1024 px), U-Net (1024x512 px), CNN (1024x512 px), and K-means clustering (1024x512 px) on representative FFT images. (*Indicates full image reconstructed from half-image)

| | Dice | IoU | Precision | Recall | Hausdorff Distance | Accuracy |
|---|---|---|---|---|---|---|
| **UNet 1024x1024px** | 0.595 | 0.429 | 0.449 | 0.917 | 265.999 | 0.998 |
| **UNet 1024x512px** | 0.835 | 0.721 | 0.723 | 0.996 | 21.707 | 0.999 |
| **CNN 1024x512px** | 0.659 | 0.496 | 0.632 | 0.746 | 118.384 | 0.999 |
| **K-means Clustering** | 0.156 | 0.0968 | 0.149 | 0.287 | 255.925 | 0.993 |

**Table S4:** Comparison of segmentation performance metrics across different models.

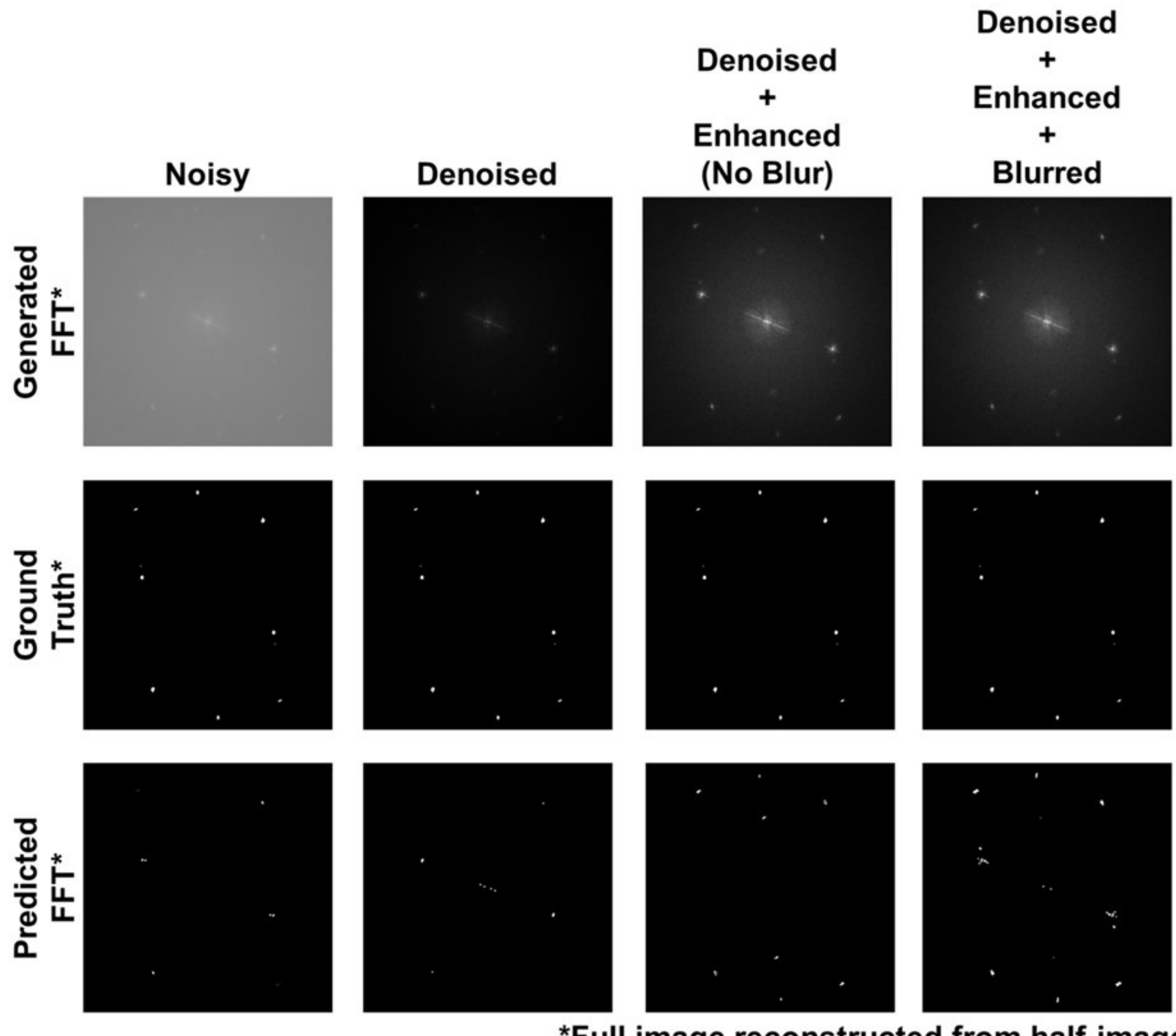

**Figure S13.** Ablation study results on FFT pre-processing

| Category | Dice Coefficient | Jaccard Score | Mean Pixel Accuracy | Hausdorff Distance |
|---|---|---|---|---|
| **Noisy** | 0.378 | 0.240 | 0.999 | 341.422 |
| **Denoised** | 0.233 | 0.142 | 0.999 | 398.855 |
| **Denoised + Enhanced (No Blur)** | 0.523 | 0.378 | 0.999 | 202.854 |
| **Denoised + Enhanced + Blurred** | 0.835 | 0.721 | 0.999 | 21.707 |

**Table S5.** Metrics of the ablation study on FFT pre-processing

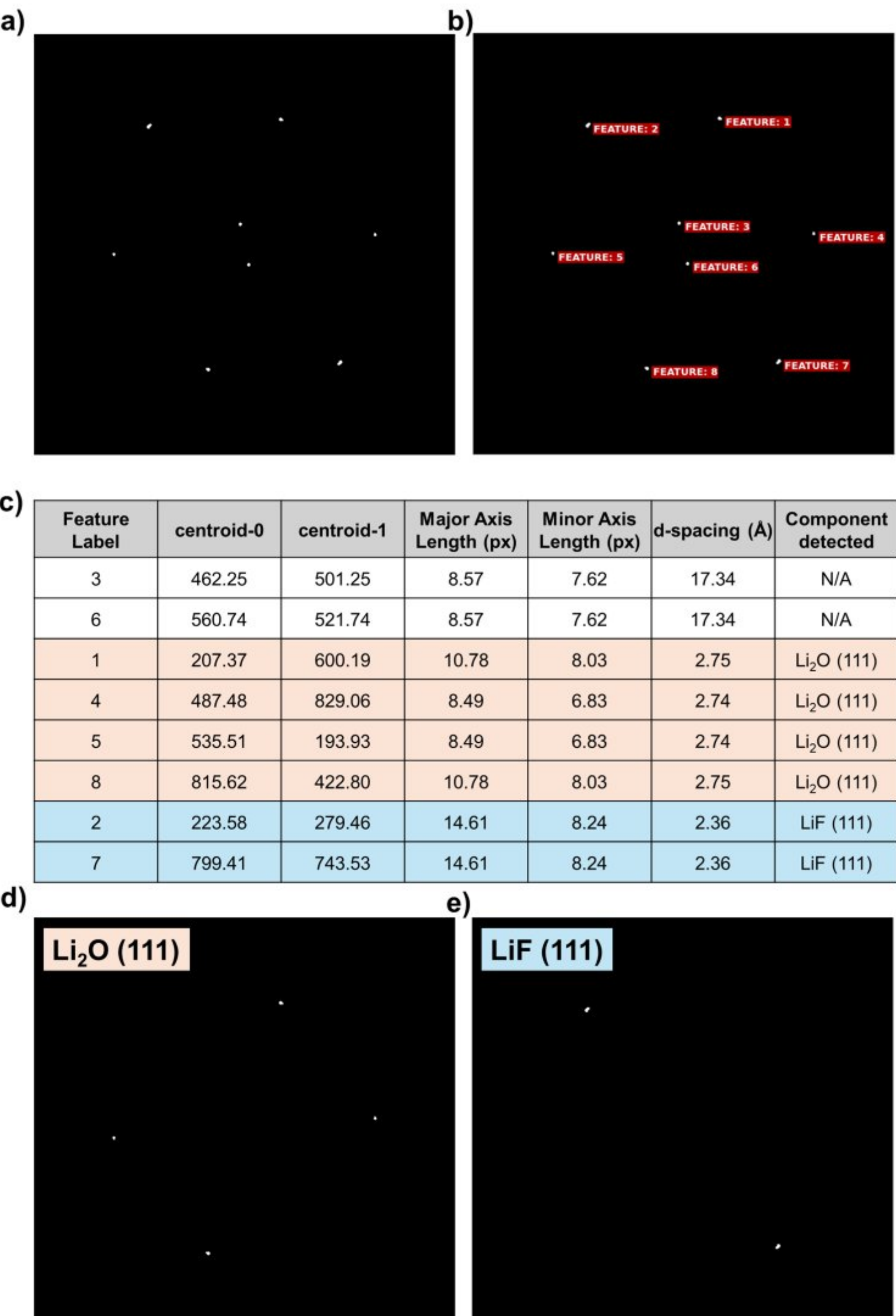

| Feature Label | centroid-0 | centroid-1 | Major Axis Length (px) | Minor Axis Length (px) | d-spacing (Å) | Component detected |
|---|---|---|---|---|---|---|
| 3 | 462.25 | 501.25 | 8.57 | 7.62 | 17.34 | N/A |
| 6 | 560.74 | 521.74 | 8.57 | 7.62 | 17.34 | N/A |
| 1 | 207.37 | 600.19 | 10.78 | 8.03 | 2.75 | $Li_2O$ (111) |
| 4 | 487.48 | 829.06 | 8.49 | 6.83 | 2.74 | $Li_2O$ (111) |
| 5 | 535.51 | 193.93 | 8.49 | 6.83 | 2.74 | $Li_2O$ (111) |
| 8 | 815.62 | 422.80 | 10.78 | 8.03 | 2.75 | $Li_2O$ (111) |
| 2 | 223.58 | 279.46 | 14.61 | 8.24 | 2.36 | LiF (111) |
| 7 | 799.41 | 743.53 | 14.61 | 8.24 | 2.36 | LiF (111) |

**Figure S14.** Component detection from the predicted FFT : **(a)** predicted FFT image **(b)** detected features using cv2.connectedComponents **(c)** Features grouped (highlighted) based on the detected components using the calculated d-spacing value. Reconstructed FFT image with only the features corresponding to component **(d)** $Li_2O$ (111) **(e)** LiF (111)

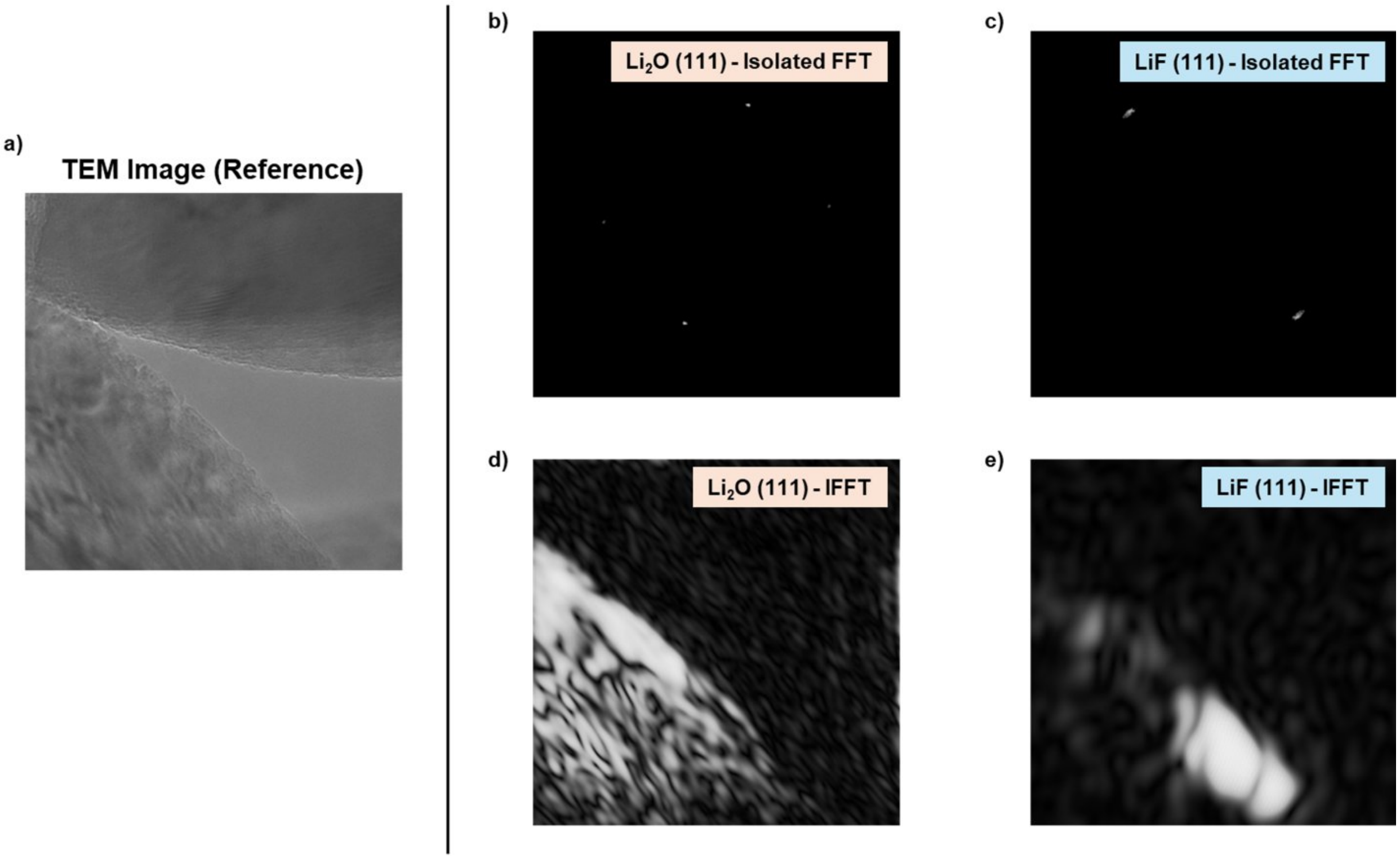

**Figure S15. (a)** Reference TEM image showing phase boundaries. **(b,c)** Isolated FFT patterns of $Li_2O$ (111) and LiF (111) planes. **(d,e)** Corresponding IFFT reconstructions revealing spatial distribution of crystallographic phases and their interfacial characteristics.

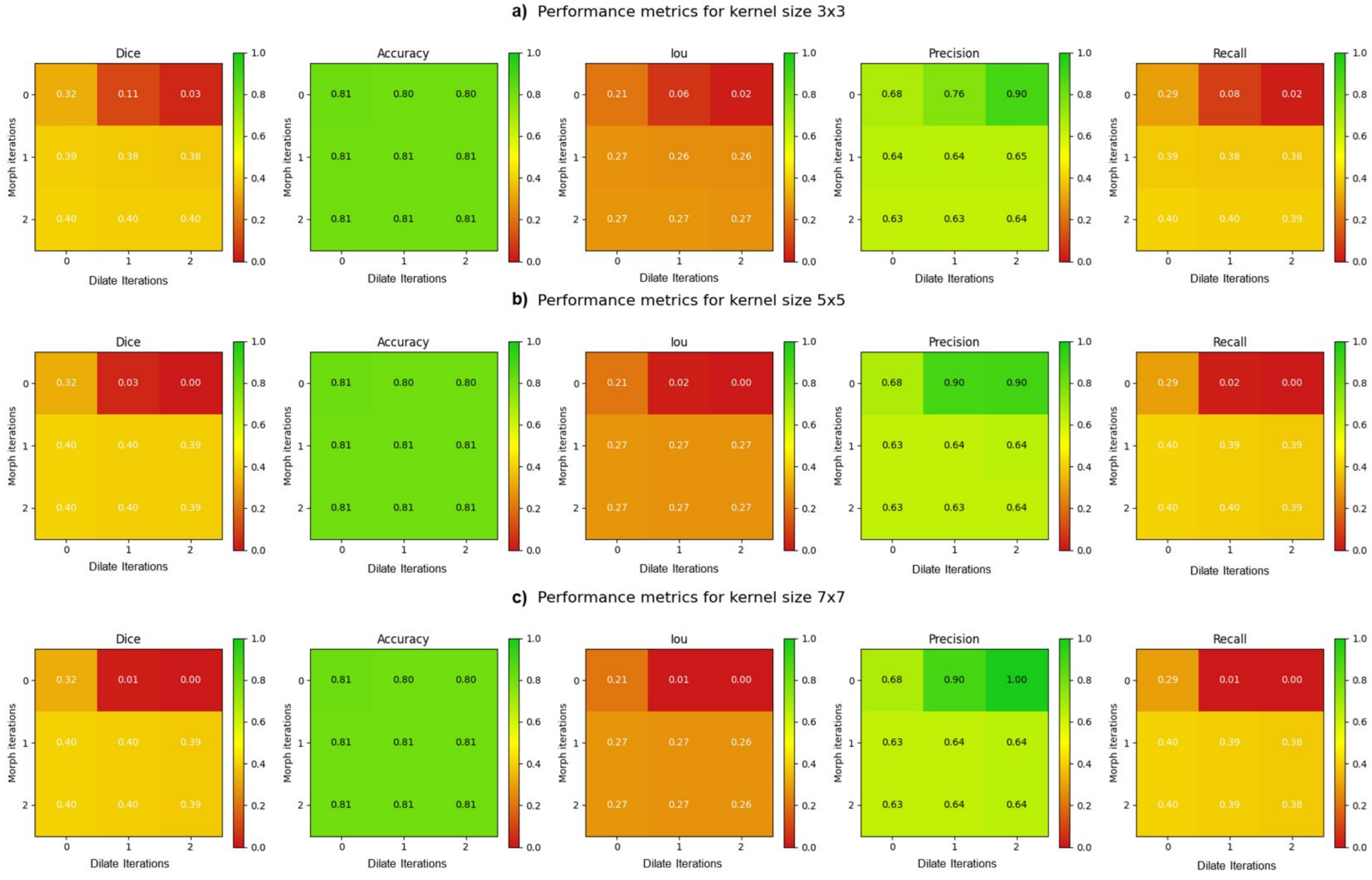

**Figure S16.** Watershed segmentation performance metrics heatmap for **(a)** 3×3, **(b)** 5×5, and **(c)** 7×7 kernel sizes.

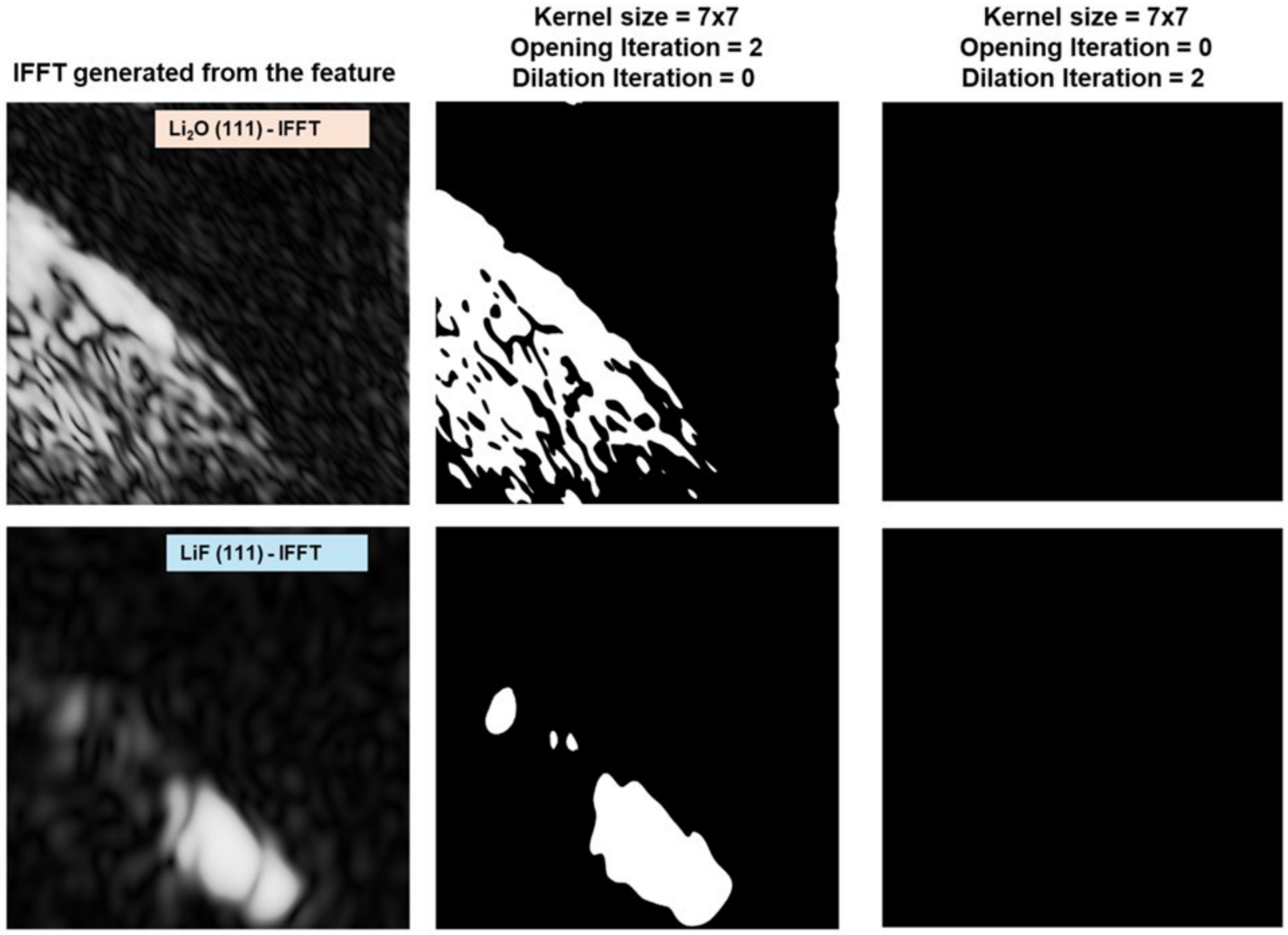

**Figure S17.** Comparison of watershed segmentation results of IFFT mapping of Li2O (111) and LiF (111) with different parameter combinations **:** 7×7 kernel size, 2 opening iterations, 0 dilation iterations; and 7×7 kernel size, 0 opening iterations, 2 dilation iterations.

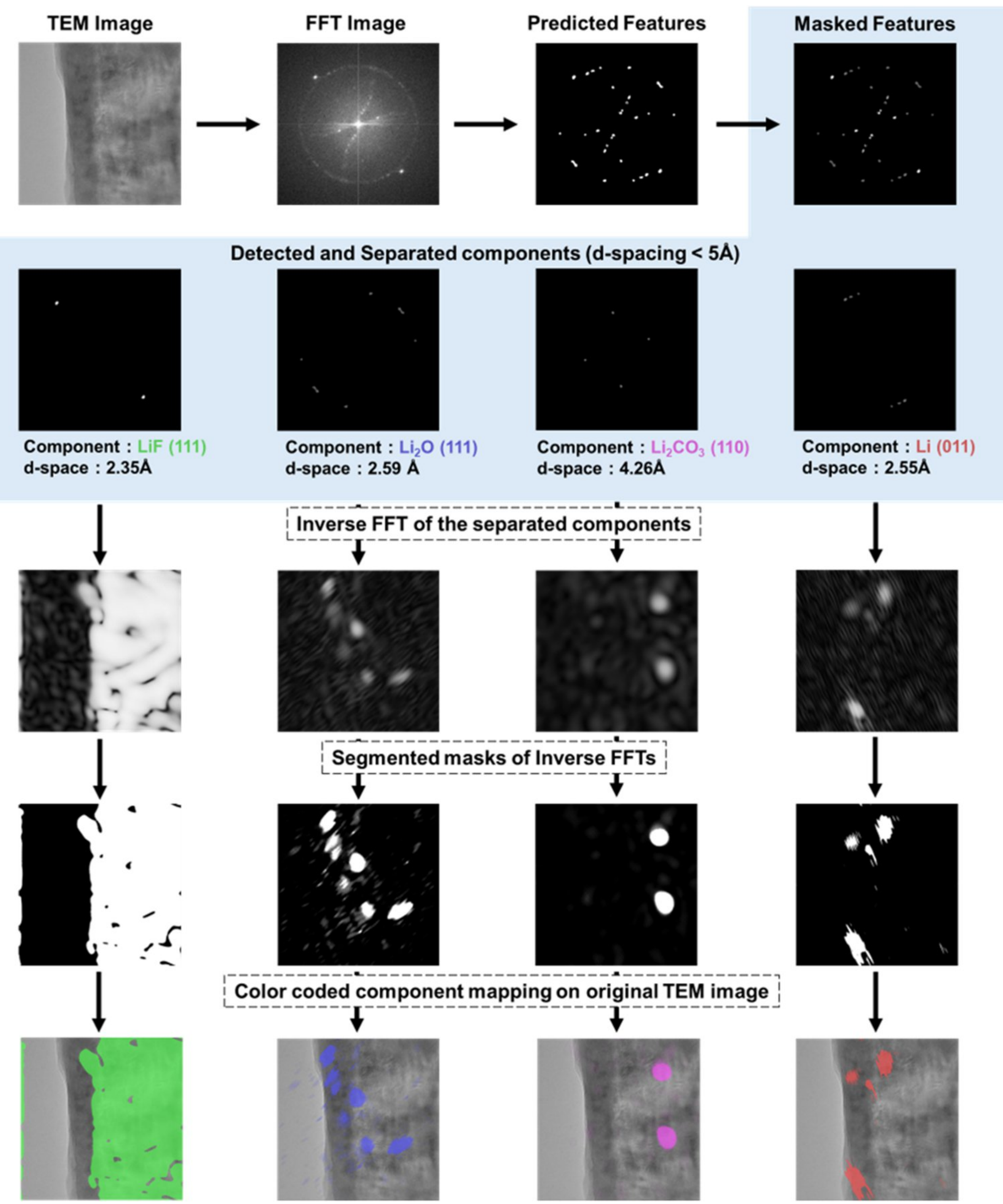

**Figure S18.** Overall program flow demonstrated with an example of a TEM image

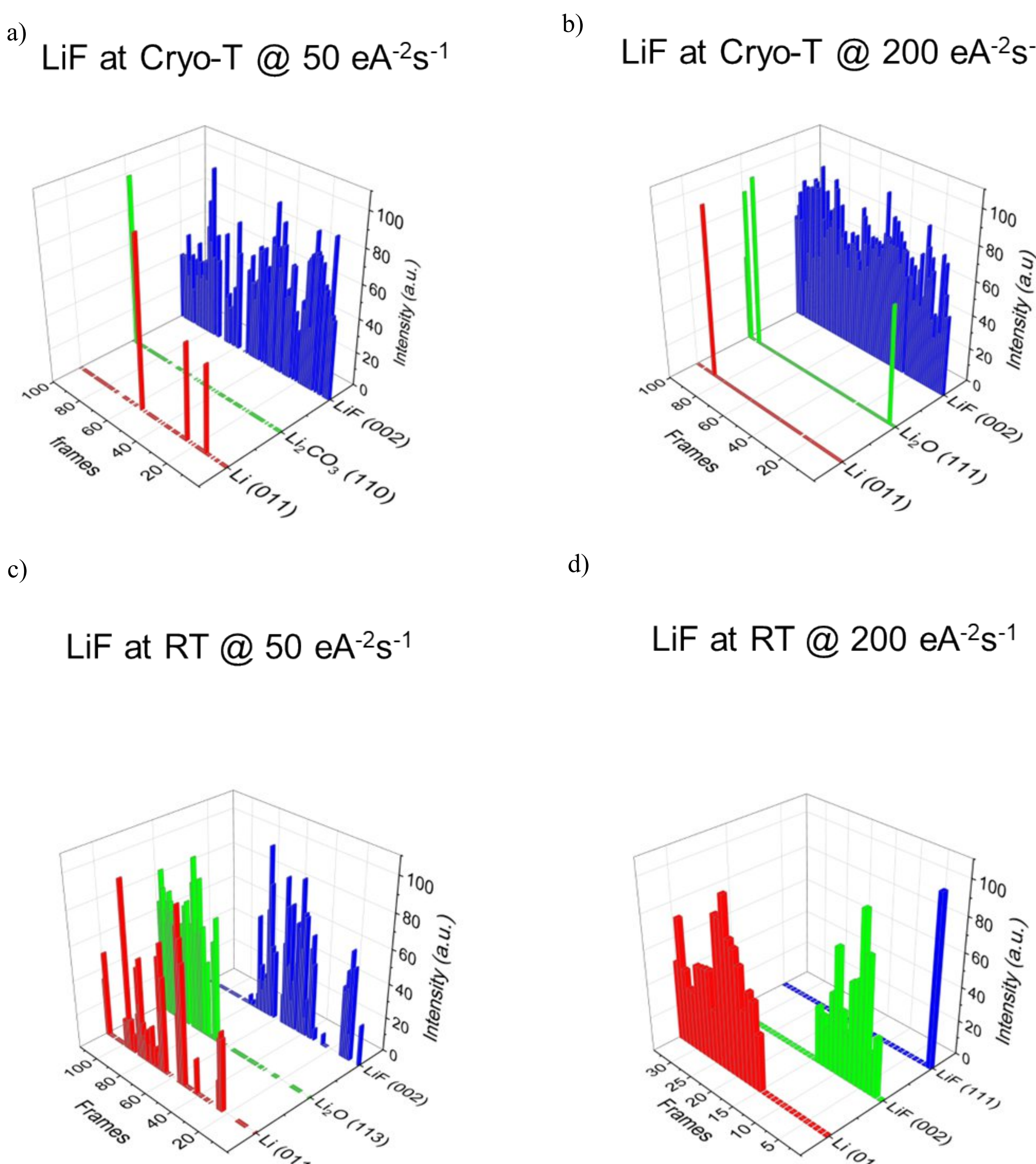

**Figure S19.** Intensity profiling of analyzed TEM images of LiF at cryo temperature at dose rate of (a) 50 e $A^{-2}$ $s^{-1}$ (b) 200 e $A^{-2}$ $s^{-1}$, and at room temperature at dose rate of (c) 50 e $A^{-2}$ $s^{-1}$. (d) 200 e $A^{-2}$ $s^{-1}$

## ■ GUI DEVELOPMENT FOR AUTOMATED TEM ANALYSIS

The creation of user-friendly interfaces is increasingly important in scientific software, especially when dealing with non-technical users. Tkinter, a popular Python library for GUI development, was utilized to create a GUI interface for automated TEM analysis in this study. The interface is designed to process TEM images and present the output in a results window. The processed images are displayed under the "SAMPLE NAME" column, and when a processed image is selected, the corresponding TEM image is displayed along with the list of detected components under the "COMPONENTS LIST" column. Each component in the list can be selected to map the region of the component on the TEM image and provide information about the percentage match with the database. This interface is expected to greatly simplify and expedite the analysis of TEM images by providing easy navigation and interactive visualizations. We employ the "multiprocessing" library to undertake the concurrent processing of multiple files. By accepting a parameter encompassing a multitude of file names, we utilize asynchronous parallel processing to expedite the execution of said files.

The developed tool offers several potential advantages over current state-of-the-art analysis programs, particularly when applied to large datasets:

**Efficient Batch Processing of TEM Images:**

The developed program significantly expedites the analysis of large TEM datasets. It can batch process up to 100 images within 30 minutes, a substantial improvement compared to manual processing using GATAN software, which typically takes around 5 minutes per image. This manual approach can become even more time-consuming depending on the complexity of the FFT images and the number of features to be identified. This program thus offers a valuable tool for researchers dealing with high-volume TEM data, accelerating their analysis workflow and enabling more efficient extraction of critical information.

**Precise and Automated Identification of Periodic Components:**

The developed program introduces a novel capability for automatically identifying periodic components in TEM images. By referencing a user-defined database, it assigns the nearest d-spacing material to each detected component in the FFT. This functionality surpasses the capabilities of manual analysis software like GATAN, which lacks automated identification. Furthermore, the high accuracy of the program in detecting periodic components is clearly demonstrated in **Table 1**, highlighting its potential to streamline and enhance the precision of TEM analysis workflows.

**Unique Intensity Distribution Profiling:**

The developed program introduces a novel and powerful feature for analyzing the evolution of periodic components in TEM images. It generates an intensity distribution profile (**Fig. 5(a)**) (**Fig. S13**) by tracking the intensity of specific features across a series of image slices. This unique capability is not available in any existing analysis software and soffers valuable insights into dynamic processes, such as beam damage in LiF samples as demonstrated in this work. This feature is particularly valuable for analyzing in-situ experiments, where large datasets need to be efficiently processed and interpreted.

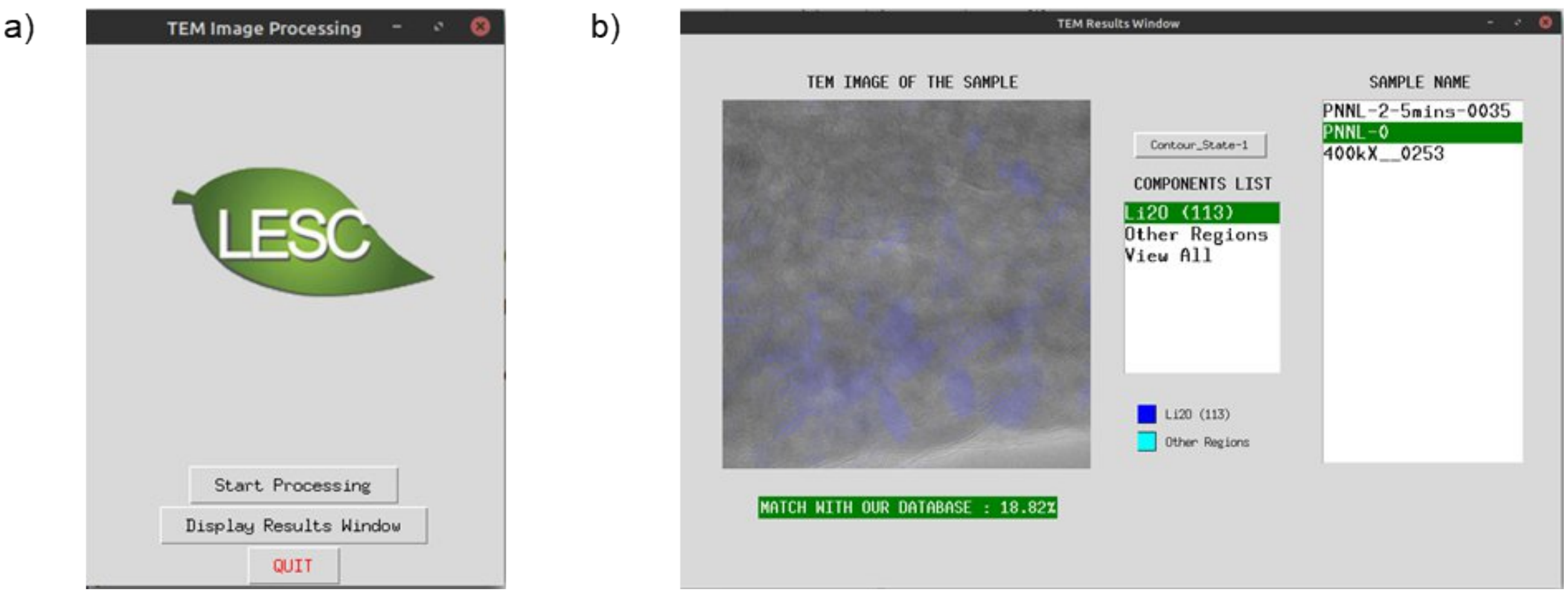

**Figure S20.** (a) TEM Processing GUI Tool start menu. (b) TEM Results window displaying corresponding components, component mapping on the TEM image.